\documentclass[twoside]{article}

\usepackage{import}

\usepackage[utf8x]{inputenc}
\usepackage{times}

\usepackage{changepage}

\usepackage{fancyhdr}
\usepackage{appendix} 

\usepackage{graphicx}
\usepackage{subcaption}

\usepackage{booktabs}
\usepackage{makecell}

\usepackage{thmtools, thm-restate}

\usepackage{algorithm,algorithmic} 

\usepackage[algo2e,inoutnumbered,algoruled,vlined]{algorithm2e}

\usepackage{amsmath,amsfonts,amssymb}
\usepackage{bm}

\usepackage{../style/math_letters} 

\usepackage{bm}
\usepackage{enumerate}
\usepackage{mathtools}

\newcommand{\BlackBox}{\rule{1.5ex}{1.5ex}}  
\newenvironment{proof}{\par\noindent{\bf Proof\ }}{\hfill\BlackBox\\[2mm]}
\newtheorem{example}{Example} 
\newtheorem{theorem}{Theorem}
\newtheorem{lemma}[theorem]{Lemma} 
 
\newtheorem{remark}[theorem]{Remark}
\newtheorem{assumption}[theorem]{Assumption}
\newtheorem{corollary}[theorem]{Corollary}
\newtheorem{definition}[theorem]{Definition}

\newenvironment{mydefinition}[1]{
	
	\begin{minipage}{0.91\textwidth}
		\vspace{2mm}	
		\begin{definition}[#1]
		}{
		\end{definition}
		\vspace{0mm}
	\end{minipage}
	
}

\newenvironment{myassumption}[1]{
	
	\begin{minipage}{0.91\textwidth}
		\vspace{2mm}	
		\begin{assumption}[#1]
		}{
		\end{assumption}
		\vspace{0mm}
	\end{minipage}
	
}

\newenvironment{mylemma}[1]{
	
	\begin{minipage}{0.91\textwidth}
		\vspace{2mm}	
		\begin{lemma}[#1]
		}{
		\end{lemma}
		\vspace{0mm}
	\end{minipage}
	
}

\newenvironment{mytheorem}[1]{
	
	\begin{minipage}{0.91\textwidth}
		\vspace{2mm}	
		\begin{theorem}[#1]
		}{
		\end{theorem}
		\vspace{0mm}
	\end{minipage}
	
}

\newenvironment{mycorollary}[1]{
	
	\begin{minipage}{0.91\textwidth}
		\vspace{2mm}	
		\begin{corollary}[#1]
		}{
		\end{corollary}
		\vspace{0mm}
	\end{minipage}
	
}

\DeclareMathOperator{\sign}{sign}

\usepackage[disable]{todonotes}

\usepackage{multicol}

\usepackage[accepted]{aistats2023}
\usepackage[round]{natbib}

\graphicspath{{figures/}}

\usetikzlibrary{arrows}

\begin{document}
	
\runningtitle{Risk-aware linear bandits with convex loss}
\runningauthor{Saux, Maillard}

%

%

\twocolumn[

\aistatstitle{Risk-aware linear bandits with convex loss}

\aistatsauthor{ Patrick Saux \And Odalric-Ambrym Maillard}

\aistatsaddress{ Inria, Univ. Lille, CNRS, Centrale Lille,\\ UMR 9198-CRIStAL, F-59000 Lille, France \And Inria, Univ. Lille, CNRS, Centrale Lille,\\ UMR 9198-CRIStAL, F-59000 Lille, France} ]

\begin{abstract}
  In decision-making problems such as the multi-armed bandit, an agent learns sequentially by optimizing a certain feedback. While the mean reward criterion has been extensively studied, other measures that reflect an aversion to adverse outcomes, such as mean-variance or conditional value-at-risk (CVaR), can be of interest for critical applications (healthcare, agriculture). Algorithms have been proposed for such risk-aware measures under bandit feedback without contextual information. In this work, we study contextual bandits where such risk measures can be elicited as linear functions of the contexts through the minimization of a convex loss. A typical example that fits within this framework is the expectile measure, which is obtained as the solution of an asymmetric least-square problem. Using the method of mixtures for supermartingales, we derive confidence sequences for the estimation of such risk measures. We then propose an optimistic UCB algorithm to learn optimal risk-aware actions, with regret guarantees similar to those of generalized linear bandits. This approach requires solving a convex problem at each round of the algorithm, which we can relax by allowing only approximated solution obtained by online gradient descent, at the cost of slightly higher regret. We conclude by evaluating the resulting algorithms on numerical experiments.
\end{abstract}

\section{INTRODUCTION}

Contextual bandits are sequential decision-making models where at each time step an agent observes a set of possible actions, or contexts, plays one of them and receives a stochastic reward, the distribution of which is a function of the selected action. The goal of the agent is to learn a policy in order to maximize rewards, facing the classical exploitation-exploration dilemma. A prominent example of such models is the linear bandit, which assumes a linear relationship actions and the \textit{mean} rewards. In this setting, a standard learning strategy consists in estimating the reward model by ridge regression coupled with an appropriate exploration scheme, e.g., optimism \citep{abbasi_yadkori2011}, Thompson sampling \citep{agrawal2013thompson, abeille2017linear} or information-directed \citep{russo2014learning, kirschner2021asymptotically}.

One limitation of this setting is that real-world agents may assess rewards with a different criterion than the mean. While mathematically convenient, the latter is known to equally weight large positive and negative outcomes, possibly leading to risky policies unsuitable to critical applications, and is also sensitive to outliers. In contrast, \textit{risk-aware} measures emphasize different characteristics of the reward distribution, e.g., by stressing out the impact of adverse outcomes \citep{dowd2007measuring}. Such measures include the mean-variance \citep{markowitz1952march}, conditional Value-at-Risk \citep{rockafellar2000optimization}, which is a special case of spectral risk measures \citep{acerbi2002spectral}, entropic risk \citep{maillard2013robust} and the expectiles \citep{newey1987asymmetric}. These measures, in particular the conditional Value-at-Risk (CVaR), have been studied as alternatives to the mean criterion in classical multi-armed bandits, that is without contextual information
\citep{galichet2013exploration, gopalan2017weighted, cassel2018general, tamkin2019distributionally, prashanth2020concentration, pandey2021estimation,  baudry2021optimal}. In distributional reinforcement learning, quantile regression has been studied for DQN \citep{dabney2017distributional}. Recently, bandits with contextual mean-variance and CVaR have been applied to vehicular communication \citep{wirth2022risk}. Despite promising empirical results, these contributions are largely devoid of theoretical regret guarantees.

In this work, we investigate an extension of the linear bandit where a given risk measure, rather than the mean, is linearly parametrized by the chosen actions. Specifically, we consider the case of convex risk measures which can be elicited as minimizers of certain loss functions, which naturally extends the standard ridge regression. This definition covers quantiles, expectiles and entropic risk, and can be extended to mean-variance and conditional value-at-risk using multivariate risk measures. To our knowledge, this setting is new, although related to existing approaches, such as bandits with regression oracles \citep{foster2020beyond} and generalized linear bandits (GLB) \citep{filippi2010parametric, li2017provably, faury2020improved}, that go beyond reward linearity while still working under the mean criterion.

\paragraph{Contributions} We introduce a generalization of LinUCB to a large class of so-called \textit{elicitable} risk measures, which includes the expectiles and the entropic risk. In contrast with the standard mean-linear bandit, learning the linear mapping between actions and risk measures cannot be performed sequentially, which presents theoretical and numerical challenges similar to GLB. We derive time-uniform confidence sets (Proposition~\ref{prop:conf}) based on the method of mixtures \citep{delapena_self_normalized} and introduce a geometric condition (Lemma~\ref{lem:transport}) to ensure sublinear regret in this new setting (Theorem~\ref{thm:regret_ucb}). Using recent developments on time-uniform matrix concentration, we further strengthen the regret bound in the case of stochastic actions with a known covariance lower bound (Theorem~\ref{thm:regret_ucb_bis}). To mitigate the numerical burden, we introduce an episodic version of LinUCB with online gradient descent approximation (Theorem~\ref{thm:regret_ucb-ogd}), inspired by previous works on online regression \citep{korda2015fast} and the recent literature on approximate Thompson sampling for GLB \citep{ding2021efficient}.

\paragraph*{Notations} We consider the Euclidean space $\left(\R^d, \langle \cdot, \cdot \rangle\right)$ and denote by $I_d$ the identity matrix of $\R^d$. For a positive definite matrix $P\in\cS^{++}_d(\R)$ and a vector $x\in\R^d$, we define the norm $\lVert x\rVert_{P}=\sqrt{\langle x, Px\rangle}$. When $P=I_d$, we let $\lVert \cdot \rVert_{P}=\lVert \cdot \rVert_2$ be the $L^2$ norm. $\cB^d_{\lVert \cdot \rVert}(x, R)$ denotes the ball in $\bR^d$ centred on $x$ of radius $R$ with respect to the norm $\lVert \cdot \rVert$. $A\preccurlyeq B$ stands $B-A\in\cS_d(\bR)$ (positive semidefinite matrix). For $K\in\N$, $[K]$ denotes the set $\lbrace 1, \dots, K\rbrace$. For a set $E$, $2^E$ denotes the set of all subsets of $E$.


\section{CONTEXTUAL BANDITS WITH RISK}

We consider the standard contextual bandit setting where an agent sequentially observes at time $t\in\bN$ a decision set $\cX_t\subseteq \bR^d$, then chooses an action $X_t\in\cX_t$ and receives a stochastic reward $Y_t$, the distribution of which is dependent on $X_t$. More formally, let $\cX=\cup_{t\in\bN} \cX_t$ and $\Phi\colon \cX \rightarrow \cP(\bR)$ a mapping from actions to reward distributions, so that the agent receives at time $t$ the reward $Y_t\sim \Phi(X_t)$. We denote by $\cF_t=\sigma\left(\cX_1, X_1, Y_1, \dots, \cX_{t-1}, X_{t-1}, Y_{t-1}, \cX_t, X_t\right)$ the $\sigma$-algebra corresponding to the information available to the agent at time $t$ (that is after choosing the action $X_t$ but before observing the reward $Y_t$). Loosely speaking, the goal of the agent is to learn a representation of the mapping $\Phi$ in order to select actions that induce high rewards. A standard inductive bias in this context is to assume a linear relation between rewards and contexts, typically of the form $Y_t=\left\langle \theta^*, X_t\right\rangle + \eta_t$, where $\eta$ is a stochastic noise process. In this work, we consider a slightly more general notion of linearity by assuming instead the existence of a factorization of the mapping between actions and rewards:
\begin{center}
\begin{tikzpicture}[node distance=0.15\textwidth]
\node (a)  {$\cX$};
\node (b) [right of=a] {$\bR^p$};
\node (c) [right of=b] {$\cP(\bR)$};

\draw[->] (a) -- node[above] {$\ell^*$} (b);
\draw[->] (b) -- node[above] {$\varphi$} (c);
\draw [->] (a) to [out=-15,in=-165] node[below] {$\Phi = \varphi\circ \ell^*$} (c);
\end{tikzpicture}
\end{center}
where $\ell^*\colon \cX\rightarrow \R^p$ denotes a linear map. In other words, the reward distribution (but not necessarily its mean) is linearly parametrized by the chosen action. We denote such a linear bandit by $(\varphi, \ell^*)$. When the distribution depends on a single parameter ($p=1$), we represent the linear form by $\ell^*\colon x\in\cX\mapsto \langle \theta^*, x\rangle$, where $\theta^*\in\R^d$, and we use the notation $(\varphi, \theta^*)$, or equivalently we say that it is represented by $Y\sim \varphi(\langle \theta^*, X\rangle)$. In the following, we also denote by $\Theta\subseteq \R^d$ parameter space.

As an example, let us consider the following Gaussian mapping $\Phi\colon x\in\cX \mapsto \mu(x) + \sigma(x) \cN(0, 1)$.
If $\sigma$ is constant and $\mu(x) = \langle \theta_{\mu}, x\rangle$, we recover a standard linear bandit model, in which the goal is to maximize the cumulative average rewards $\sum_{t=1}^{T}\mu(X_t)$. However, in many applications, the agent may be averse to high reward volatility, which can be encoded by $\mu(x) - \lambda \sigma(x)$ for some $\lambda>0$. We detail in Appendix~\ref{app:elicitable} how many standard risk measures (entropic, $p$-expectile) realize this mean-variance tradeoff.

%
%

\subsection{Overview of Risk Measures}

\paragraph{Convex Loss}
In the bandit setting, the agent faces the classical dilemma between exploitation (playing the most promising actions) and exploration (playing other actions to gain information). In most algorithms, the exploitation takes the form of a supervised estimation that consists in learning the mapping $\varphi$ at time $t$ from the past observations $\lbrace (X_s, Y_s),\ 1\leq s \leq t\!-\!1\rbrace$. When the expected reward is parametrized by $Y_t = \langle \theta^*, X_t\rangle + \eta_t$ with $\bE[\eta_t\lvert \cF_t]=0$, a standard strategy consists in estimating $\theta^*$ by ridge regression, that is $\min_{\theta\in\Theta}\sum_{s=1}^{t-1} \left(Y_s - \langle \theta, X_s\rangle\right)^2 + \frac{\alpha}{2}\lVert \theta\rVert^2_2$, where $\alpha>0$ is a regularization parameter. Assuming for now the solution is in the interior of $\Theta$, the solution can be written as $\wh{\theta}_t=\left(V^{\alpha}_t\right)^{-1}\sum_{s=1}^{t-1}Y_sX_s$, where we define the $d\times d$ positive definite matrix $V^{\alpha}_t \coloneqq \sum_{s=1}^{t-1} X_s X_s^\top + \alpha I_d$. This method presents several advantages: it can be computed efficiently via sequential matrix inversion (with complexity $\cO(d^2)$ at each step thanks to the Sherman-Morrison formula for the rank-one update $V^\alpha_{t+1} = V^\alpha_t + X_t X_t^\top$ starting from $V^\alpha_0 = \alpha I_d$) and explicit confidence ellipsoids for $\theta^*$ can be constructed analytically around $\wh{\theta}_t$ to tune exploration \citep{abbasi_yadkori2011}. The implicit limitation of this procedure is that it can only estimate the expectation $\bE[Y] = \argmin_{\theta\in\Theta} \bE[(Y-\langle\theta, X\rangle)^2]$. We call this standard setting the \textit{mean-linear} bandit.

As motivated by the example above, we aim to estimate other statistics than the mean of the reward distribution. Drawing inspiration from this simple case, we consider an arbitrary convex loss function $\cL \colon \R\times\R^p \rightarrow \R_+$ and define the \textit{risk measure} associated with loss $\cL$ for a distribution $\nu$ over $\R$ as $\rho_{\cL}(\nu) = \arg\min_{\xi\in \R^p}\bE_{Y\sim\nu}\left[ \cL(Y, \xi)\right]$. We assume here that the $\argmin$ is unique for simplicity (which is the case if $\cL$ is strongly convex) and will sometimes use the notation $\rho_{\cL}(Y)$ instead of $\rho_{\cL}(\nu)$ for a random variable $Y$ with distribution $\nu$. Similarly, we define the conditional risk measure $\rho_{\cL}(\nu\lvert \cG) = \arg\min_{\xi\in \R^p}\bE_{Y\sim\nu}\left[ \cL(Y, \xi)\lvert \cG\right]$ for any event $\cG$ with positive measure. Note that with this definition, $\rho_{\cL}(\nu)$ is a vector in $\R^p$. When $p=1$, we call these \textit{scalar} risk measures. The motivation to consider vector-valued risk measures comes from the fact that not every measure of interest can be \textit{elicited} as a scalar risk measure, which we develop in the next paragraph. 

\paragraph{Elicitable Risk Measure} Scalar risk measures that can be expressed as minimizers of such loss functions are known as  (first-order) \textit{elicitable} risk measures \citep{ziegel2016coherence}. Examples of such measures include the mean, the median, and more generally any quantile and expectile, which we further discuss below as special cases of risk measures associated with convex potentials. Other examples are any generalized moments $\rho(\nu)=\bE_{Y\sim\nu}[T(Y)]$, where $T\colon\R\rightarrow\R$ is a $\nu$-integrable mapping, and the entropic risk defined by $\rho_{\cL}(\nu) = \frac{1}{\gamma}\log\bE_{Y\sim\nu}[e^{\gamma Y}]$ \citep{maillard2013robust}. Unfortunately, not all measures commonly encountered in the risk literature are first-order elicitable. In particular, neither the variance nor the CVaR can be expressed as scalar risk measures with respect to a convex loss \citep{fissler2015expected, fissler2016higher}. However, they are second-order elicitable, in the sense that the pairs (mean, variance) and (VaR, CVaR) are jointly elicitable. We refer to Appendix~\ref{app:elicitable} for a summary and further interpretation of elicitable risk measures.

We say that the loss $\cL$ is adapted to the linear bandit $(\varphi, \ell^*)$ if for all $x\in\cX$, $\ell^*$ is a minimizer of $\bE\left[ \cL\left(Y, \ell(x)\right) \right]$ among all linear forms $\ell\colon \cX\rightarrow \R^p$, where $Y\sim\varphi\circ \ell^*(x)$ denotes the reward random variable of the linear bandit when action $x$ is played. Intuitively, this means that the risk measure $\rho_{\cL}$ of the reward distribution is linearly parametrized by the actions, the same way the expected reward is a linear form of the action in the standard mean-linear bandit.

\begin{remark}
	The above definition is written in the general case of a vector-valued risk measure $\rho_{\cL}$. In the rest of this paper, we only consider scalar risk measure and leave the extension to measures like CVaR for further work. We say that $\cL$ is adapted to the linear bandit $(\varphi, \theta^*)$ if for all
	$x\in\cX$, we have that $\theta^* \in \arg\min_{\theta\in\Theta} \bE_{Y\sim \varphi(\langle\theta^*, x\rangle)}\left[ \cL\left(Y, \langle \theta, x\rangle\right) \right]$.
\end{remark}

\paragraph{Convex Potential} A special case of interest is when the convex loss $\cL=\cL_{\psi}$ derives from a potential $\psi$, that is when $\cL_{\psi}(y,\xi) = \psi(y\!-\!\xi)$. This includes the ordinary least square potential associated to the mean, as well as quantiles and expectiles. We assume the reader to be familiar with the former, but perhaps less so with the latter. Following \citet{newey1987asymmetric}, we define the $p$-expectile of $\nu$ for $p\in(0, 1)$ as $\argmin_{\xi\in\R} \bE_{Y\sim\nu}[\lvert p - \ind_{Y<\xi}\rvert (Y-\xi)^2]$. Expectiles have been studied in particular in the context of risk management \citep{bellini2017risk} and risk-aware Bayesian optimization \citep{torossian2020bayesian}. Furthermore, under some symmetry conditions, quantiles and expectiles are known to coincide \citep{abdous1995relating}, and thus expectiles can be seen as a smooth (in particular differentiable) generalization of quantiles (see \cite{philipps2022interpreting} for further interpretation of the notion of expectiles). We refer the reader to Table~\ref{table:potential} for a summary of risk measures elicited by convex potentials. 

	\begin{table}
		\caption{Example of Risk Measures Elicited by Convex Potentials.}
		\label{table:potential}
		\centering
		\begin{tabular}{lll}
			\toprule
			Name     & Potential $\psi(z)$     & Risk measure $\rho_{\psi}$ \\
			\midrule
			Mean & $z^2/2$  & $\rho_{\psi} = \int y\nu(dy)$\label{eqn:OLS_potential}\\[1.0ex]
			\makecell[l]{Quantile\\ $p\in(0,1)$} & $(p-\ind_{z<0})z$  & $\int_{-\infty}^{\rho_{\psi}} \nu(dy) = p$\label{eqn:quantile_potential}\\[1.0ex]
			\makecell[l]{Expectile\\ $p\in(0,1)$} & $\lvert p-\ind_{z<0}\rvert z^2$  &\makecell[l]{ $(1\!-\!p)\int_{-\infty}^{\rho_{\psi}} \lvert y\!-\!\rho_{\psi}\rvert \nu(dy)$ \\ $= p\int_{\rho_{\psi}}^{\infty} \lvert y-\rho_{\psi}\rvert \nu(dy)$} \label{eqn:expectile_potential}\\[1.0ex]
			\bottomrule
		\end{tabular}
		\vskip -5mm
	\end{table}

In the terminology defined above, the ordinary least square potential is adapted to the mean-linear reward model $Y_t=\langle \theta^*, X_t\rangle + \eta_t$ with $\bE[\eta_t\lvert \cF_t]=0$. More generally, such an additive decomposition exists for losses derived from potentials (see Lemma~\ref{lemma:linearity_psi} in Appendix~\ref{app:convex_potentials}).

\paragraph{Non-unicity of Adapted Loss} In general, a risk measure can be described by multiple different adapted losses. First, the set of losses that elicit a given risk measure is a cone invariant by scalar translation, i.e., $\rho_{\alpha\cL + \beta} = \rho_{\cL}$ for all $\alpha>0$ and $\beta\in\R$. Other less trivial examples of non-unicity arise even for the simple mean criterion. Theorem~1 in \cite{banerjee2005optimality} shows that $\bE_{Y\sim\nu}\left[Y\right]=\rho_{\cB_{\psi}}(\nu)$ where $\psi$ is any strictly convex, differentiable function and $\cB_{\psi}\colon (y, \xi)\mapsto \psi(y) - \psi(\xi) - \psi'(\xi)(y-\xi)$ is the Bregman divergence induced by $\psi$, which generalizes the quadratic potential. In fact, every continuously differentiable loss that elicits the mean has this form (Theorem~3 and 4 in \citet{banerjee2005optimality}). Similarly, the pairs (mean, variance) and (VaR, CVaR) can be elicited by families indexed by differentiable, strictly convex functions (Table~\ref{table:elicitable}, Appendix~\ref{app:elicitable}).

\subsection{Contextual Bandits with Elicitable Risk Measures}
\paragraph{Regret} For a linear bandit $(\varphi, \theta^*)$, we define the pseudo-regret associated to a risk measure $\rho_{\cL}$ and a sequence of actions $\left(X_t\right)_{1\leq t\leq T}$ as $\cR_T\!=\!\sum_{t=1}^T \rho_{\cL}\left(\varphi(\langle \theta^*, X^*_t\rangle)\right)\!-\!\rho_{\cL}\left(\varphi(\langle \theta^*, X_t\rangle)\right)$, where $X^*_t = \argmax_{x\in\cX_t} \rho_{\cL}\left(\varphi(\langle \theta^*, x\rangle)\right)$ is the optimal action w.r.t the risk measure $\rho_{\cL}$. By definition, if the loss $\cL$ is adapted to the linear bandit, this notion of regret reduces to $\cR_T = \sum_{t=1}^T \langle \theta^*, X^*_t\rangle - \langle \theta^*, X_t\rangle$, which is formally the same as the standard regret for mean-linear bandits. What differs though is the meaning of $\langle \theta^*, X_t\rangle$, which now represents an elicitable risk measure for the reward distribution. As an example, this paves a way for expectile-linear bandit of the form $Y_t = \langle \theta^*, X_t\rangle + \eta_t$ where the conditional expectile of $\eta_t$ is zero and expectile rewards are measured as linear forms of the actions $\langle \theta^*, X_t\rangle$.

\paragraph{Supervised Estimation of $\theta^*$} Similarly to how ridge regression provides natural estimators of the mean, we define $\wh{\theta}_t \in \argmin_{\theta\in\Theta} \sum_{s=1}^{t-1} \cL(Y_s, \langle \theta, X_s\rangle) + \frac{\alpha}{2}\lVert \theta \rVert^2_2$, which corresponds to the empirical risk minimization associated to loss $\cL$, with $L^2$ regularization parameter $\alpha>0$. Assuming that $\cL$ is differentiable and that $\wh{\theta}_t$ is in the interior of $\Theta$, this estimator is characterized by the equation $\alpha \wh{\theta}_t\!=\!-\sum_{s=1}^{t-1} \partial \cL(Y_s, \langle \wh{\theta}_t, X_s\rangle)X_s$, where $\partial\cL(y, \xi)$ stands for the derivative of $\xi\!\mapsto\!\cL(y, \xi)$. When $\wh{\theta}_t$ is not in the interior of $\Theta$, an additional projection onto $\Theta$ is necessary, which we denote by the operator $\Pi$ (such an operator is detailed in Section~\ref{subsec:martingale}). We also define $H^{\alpha}_t(\theta) = \sum_{s=1}^{t-1}\partial^2 \cL\left(Y_s, \langle \theta, X_s\rangle\right)X_sX_s^\top\!+\!\alpha I_d$ the Hessian of the empirical loss at $\theta$ of the minimization problem.


We note that when $\cL$ derives from the quadratic potential $\psi(\xi) = \xi^2 /2$, it holds that $H^\alpha_t(\theta) = V^{\alpha}_t$ and we thus fall back to the mean-linear case. For all other choices of the loss function $\cL$, the Hessian $H^\alpha_t$ depends on $\theta$, and in particular no closed-form expression of $\wh{\theta}_t$ in terms of the inverse of $H^\alpha_t$ is available. As we detail in the next sections, this introduces technical challenges to the analysis of linear bandit algorithms and forces the use of convex programming algorithms to numerically evaluate $\wh{\theta}_t$. 

\begin{remark}
	Similar complications arise in the case of generalized linear bandits (GLB) $Y_t = \mu(\langle \theta^*, X_t\rangle) + \eta_t$, with $\bE[\eta_t\lvert \cF_t]=0$ and $\mu$ a nonlinear link function. Under parametric assumptions on $Y_t$ (typically one-dimensional exponential family), GLB can be seen as a special case of the risk-aware setting with $\cL$ the negative log-likelihood loss, with the analogy $\mu\leftrightarrow \partial\cL$. Despite this formal similarity, GLB is designed solely to optimize the mean criterion. Another difference with our setting is that regret for GLB is commonly defined as $\sum_{t=1}^T\mu(\langle \theta^*, X^*_t\rangle) - \mu(\langle \theta^*, X_t\rangle)$, which is smaller than $\cR_T$ when $\mu$ is contracting.
	
\end{remark}

\paragraph{Extension of LinUCB to Convex Losses} The main benefit of the formulation of risk-awareness in terms of convex losses is that it suggests a transparent generalization of the standard LinUCB algorithm (OFUL in \citet{abbasi_yadkori2011}, Ch.19 in \citet{lattimore2020bandit}), essentially substituting the least-squares estimate with the empirical risk minimizer associated with $\cL$. The general idea of such optimistic algorithms is to play at time $t$ the action $x\in\cX_t$ with the highest plausible reward. In the mean-linear case with ridge regression, this highest plausible reward takes the form of $\langle\wh{\theta}_t, x\rangle + \gamma_t(x)$, where $\gamma_t(x)$ is a certain action-dependent quantity also known as the \textit{exploration bonus}. We write the general structure of our extension of LinUCB (CR for Convex Risk)  in Algorithm~\ref{algo::UCB_convex_risk}.
 
\begin{algorithm}[]
	\caption{LinUCB-CR \label{algo::UCB_convex_risk}}
	\SetKwInput{KwData}{Input} 
	\KwData{regularisation parameter $\alpha$, projection $\Pi$,\\ exploration bonus sequence $(\gamma_t)_{t\in\N}$.}
	\SetKwInput{KwResult}{Initialization}
	\SetKwComment{Comment}{$\triangleright$\ }{}
	\SetKwComment{Titleblock}{// }{}
	\KwResult{Observe $\cX_1$.} 
	\For{$t=1, \dots, T$}{
		$\widehat{\theta}_t \in \arg\min_{\R^d} \sum_{s=1}^{t-1} \cL(Y_s, \langle \theta, X_s\rangle) + \frac{\alpha}{2}\lVert \theta \rVert^2_2$ \Comment*[r]{Empirical risk minimization}
		$\bar{\theta}_t = \Pi(\wh{\theta}_t)$ \Comment*[r]{Projection}
		$X_t = \arg\max_{x\in\cX_t} \langle\bar{\theta}_t, x\rangle + \gamma_t(x)$ \Comment*[r]{Play arm}
		Observe $Y_t$ and $\cX_{t+1}$.
	}
\end{algorithm}

\section{ANALYSIS OF LinUCB-CR}

The goal of this section is to derive an exploration bonus sequence $(\gamma_t)_{t\in\N}$ and a projection operator $\Pi¸$ that ensure sub-linear regret of the corresponding LinUCB instance. To this end, we introduce the following control on the curvature of the adapted loss $\cL$.

\begin{restatable}[Bounded Loss Curvature]{ass}{assumptionlosscurvature}\label{ass:loss_curvature}
There exists $m$ and $M$ such that
\[\forall y, \xi\in\R,\ m\leq \partial^2\cL(y, \xi)\leq M\,.\]
We call the parameter $\kappa=\frac{M}{m}$ the \textit{conditioning} of $\cL$.
\end{restatable}

\begin{remark} This assumption is reminiscent of the standard lower bound on the derivative of the link function $\mu'$ commonly encountered in the GLB literature.
\end{remark}

\subsection{Martingale Property and Concentration}\label{subsec:martingale}
A key property for the analysis of mean-linear bandits is that the sum process $\sum_{s=1}^{t-1} \eta_s X_s$ naturally defines a vector-valued martingale in $\R^d$ with respect to the filtration $\cF_t$ \citep{abbasi_yadkori2011}. This is not the case in general for bandits associated with generic convex losses. Instead, for a given loss $\cL$, we know that $\theta^* = \argmin_{\theta}\bE[\cL(Y_t, \langle\theta, X_t\rangle)\lvert\cF_t]$. Assuming $\cL$ is differentiable and using the shorthand $\partial^j \cL^*_t=\partial^j \cL(Y_t, \langle \theta^*, X_t\rangle)$ for $j\in\bN$ and $t\in\bN$, this implies $\bE[\partial^1\cL^*_t\lvert \cF_t]=0$ since $X_t$ is measurable with respect to $\cF_t$. A direct consequence of this is that $S_t=\sum_{s=1}^{t-1} \partial^1\cL^*_s X_s$ defines a $\cF$-martingale. This process is at the heart of the next proposition, which establishes confidence bounds using the method of mixture (see \citet{delapena_self_normalized} in general and \citet{abbasi_yadkori2011, faury2020improved} for specific applications to contextual bandits). To this end, we detail below a helpful transformation of the sum process $S$ into a nonnegative supermartingales (Lemma~\ref{lem:supermart}) under a standard sub-Gaussian assumption (Assumption~\ref{ass:subgauss}) and state the high-probability uniform deviation bound we obtain (Proposition~\ref{prop:conf}), the proof of which is deferred to Appendix~\ref{app:proof_conf}.

\begin{restatable}[Sub-Gaussian]{ass}{asssubgauss}\label{ass:subgauss}
	$\partial^1 \cL^*$ is a conditionally sub-Gaussian process, i.e., there exists $R>0$ such that
	\[\forall t\in\bN,\ \forall \lambda\in\bR,\ \log\bE\left[ \exp\left(\lambda \partial^1 \cL^*_t\right) \lvert \cF_t \right] \leq \frac{\lambda^2 R^2}{2}\,.\]
\end{restatable}

\begin{restatable}[Supermartingale Control]{lem}{lemsupermart}\label{lem:supermart} Under Assumptions~\ref{ass:loss_curvature}-\ref{ass:subgauss}, there exists $\sigma>0$ such that for any $t\in\N$ and $\lambda\in\R^d$, the following holds: 
\[\bE\left[\exp\left(\langle\lambda, X_t\rangle \partial^1\cL^*_t - \frac{\sigma^2}{2} \langle\lambda, X_t\rangle^2\partial^2\cL^*_t \right)\middle| \cF_t\right] \leq 1\,.\]
\end{restatable}

\begin{restatable}[Method of Mixtures with Convex Loss]{prop}{proptwo}\label{prop:conf}
	Let $\beta>0$. Under Assumptions~\ref{ass:loss_curvature}-\ref{ass:subgauss}, with probability at least $1 - \delta$, for  all $t\in\N$, it holds that
	\[\lVert S_t \rVert^2_{H^\beta_t(\theta^*)^{-1}} \leq \sigma^2\left(2\log\frac{1}{\delta} + \log\frac{\det H^\beta_t(\theta^*)}{\det \beta I_d}\right)\,.\]
\end{restatable}

\paragraph{Discussion on Lemma~\ref{lem:supermart} and Assumption~\ref{ass:subgauss}}

As shown in the proof, Lemma~\ref{lem:supermart} alone implies Proposition~\ref{prop:conf}. While this lemma may be valid in more general settings, we show in Appendix~\ref{app:proof_conf} how it is conveniently implied by Assumption~\ref{ass:loss_curvature} and the sub-Gaussian control of Assumption~\ref{ass:subgauss}. In the rest of the paper, in particular in the regret bounds of Theorem~\ref{thm:regret_ucb}, \ref{thm:regret_ucb_bis} and \ref{thm:regret_ucb-ogd}, $\sigma$ will refer to the parameter that appears in the supermartingale control of Lemma~\ref{lem:supermart}.


Regarding Assumption~\ref{ass:subgauss}, note that for a mean-linear bandit $Y_t = \langle \theta^*, X_t\rangle + \eta_t$ with adapted loss $\cL(y, \xi) = \frac{1}{2}(y-\xi)^2$, we have that $\partial\cL^*_t = \eta_t$, which is classically assumed to be sub-Gaussian. For other bandits, and thus other adapted losses, it may be more convenient to make assumptions on the distribution of observable quantities such as $X_t$ and $Y_t$ rather than directly on $\partial\cL^*_t$. Formally, this raises the question of how the sub-Gaussian property of a random variable $Z$ transfers to $f(Z)$ for a given mapping $f$. While to our knowledge no complete answer is available, several partial results are available in the concentration literature.
\\
\hspace*{2mm} \textit{(i)} If $Z$ is Gaussian with variance $\sigma^2$ and $f$ is $M$-Lipschitz, the Tsirelson-Ibragimov-Sudakov inequality \citep[Theorem~5.5]{boucheron2013concentration} shows that $f(Z)$ is $M\sigma$-sub-Gaussian. In particular, the Lipschitz assumption holds for $\partial\cL$ if the loss curvature is bounded from above by $M$. More generally, if $Z$ can be written as a $\sigma$-Lipschitz function of a $\cN(0, 1)$, then $f(Z)$ is $M\sigma$-sub-Gaussian.\\
\hspace*{2mm}\textit{(ii)} If the density of $Z$ is strongly log-concave, then $f(Z)$ is sub-Gaussian (with parameter related to the largest eigenvalue of the Hessian of the log-density, see \citet[Theorem~5.2.15]{vershynin2018high}).\\
\hspace*{2mm}\textit{(iii)} If $Z$ is bounded (i.e., actions and rewards are bounded) and $f$ is Lipschitz and separately convex, then $f(Z)$ is sub-Gaussian (application of the entropy method, see e.g.,  \citet[Theorem~6.10]{boucheron2013concentration}). The boundedness assumption can be lifted at the cost of a slightly more stringent condition than the sub-Gaussianity of $Z$, see \citet[Theorem~3]{adamczak2005logarithmic}.

In short, Assumption~\ref{ass:subgauss} holds in a variety of settings, under rather mild assumptions on either $X_t$ and $Y_t$ or the loss $\cL$. Also of note, $\partial^1\cL$ is $M$-Lipschitz under Assumption~\ref{ass:loss_curvature}.

\paragraph{Confidence Set for $\theta^*$} To help write the above confidence set in terms of $\theta^*$ and the empirical estimator $\wh{\theta}_t$, we introduce the function $F^{\alpha}_t\colon \theta\in\Theta \mapsto \sum_{s=1}^{t-1}\partial\cL\left(Y_s, \langle\theta, X_s\rangle\right)X_s + \alpha\theta\in\R^d$.
As seen above, $F^{\alpha}_t(\wh{\theta}_t)=0$ and $F^{\alpha}_t(\theta^*)=S_t+\alpha\theta^*$. Noticing that $\lVert F^{\alpha}_t(\theta^*) - F^{\alpha}_t(\wh{\theta}_t)\rVert^2_{H^\beta_t(\theta^*)^{-1}} = \lVert S_t + \alpha\theta^*\rVert_{H^\beta_t(\theta^*)^{-1}} \leq \lVert S_t\rVert_{H^\beta_t(\theta^*)^{-1}} + \alpha\lVert \theta^*\rVert_{H^\beta_t(\theta^*)^{-1}}$, we immediately derive the following result (note the use of a priori different regularization parameters $\alpha$ and $\beta$, which we exploit in the sequel).

\begin{restatable}[]{cor}{corrone}\label{corr:conf}
For $t\in\N$, $\delta\in(0,1)$, $\alpha, \beta>0$, let 
\begin{align*}
\wh{\Theta}^\delta_t = &\bigg\lbrace \theta\in\Theta,\ \lVert F^{\alpha}_t(\theta) - F^{\alpha}_t(\wh{\theta}_t)\rVert_{H^\beta_t(\theta)^{-1}} \\ 
&\leq \sigma\sqrt{2\log\frac{1}{\delta} + \log\frac{\det H^\beta_t(\theta)}{\det \beta I_d}} + \alpha\lVert \theta\rVert_{H^\beta_t(\theta)^{-1}} \bigg\rbrace\,.
\end{align*}
Then under Assumptions~\ref{ass:loss_curvature}-\ref{ass:subgauss}, it holds that 
\[
\bP\left(\forall t\in\N,\ \theta^* \in \wh{\Theta}^\delta_t\right) \geq 1 - \delta\,.
\]
\end{restatable}

We constantly use this result in the following, in particular to construct the projection operator $\Pi$. Indeed, if we define $\bar{\theta}_t\coloneqq\Pi(\wh{\theta}_t)$ as  $\argmin_{\theta\in\Theta} \lVert F^{\alpha}_t(\theta) - F^{\alpha}_t(\wh{\theta}_t)\rVert_{H^\beta_t(\theta)^{-1}}$, we have the property that $\Pi(\wh{\theta}_t)\in\wh{\Theta}^\delta_t$ with high probability.

\begin{remark}\label{rmk:local}
	Although we formulated the bounded curvature condition (Assumption~\ref{ass:loss_curvature}) globally, we note that we only require it to hold in a convex neighborhood of $\theta^*$ containing $\bar{\theta}_t$, and Corollary~\ref{corr:conf} shows that with high probability, $\lVert \theta^* - \bar{\theta}_t \rVert_2$ is bounded (going from the $H^{\beta}_t(\theta^*)^{-1}$ norm to the Euclidean norm can be done by simple positive definite matrix inequalities). Therefore, one could instead assume a \textit{local} curvature control on $\partial\cL(y, \langle \theta, x\rangle)$ for $x\in\cX_t$ and $\theta$ in a ball around $\theta^*$, in the same spirit as Assumption~1 in \citet{li2017provably} for GLB.
\end{remark}
%
%
\subsection{Optimism and Local Metrics}\label{subsec:optimism_local_metrics}

We recall here the principle of optimism in the face of uncertainty and adapt it to the framework of elicitable risk measures. We denote by $r_t = \langle \theta^*, X^*_t\rangle - \langle \theta^*, X_t\rangle$ the instantaneous regret, where $\langle \theta^*, X^*_t\rangle = \max_{x\in\cX_t} \langle \theta^*, x\rangle$ is the optimal risk measure associated with $\cL$ at time for the actions available at time $t$. Then, simple algebra shows that
\begin{align*}
	r_t 
	&= \langle \theta^* - \bar{\theta}_t, X^*_t\rangle - \langle \theta^* - \bar{\theta}_t, X_t\rangle + \langle \bar{\theta}_t, X^*_t - X_t\rangle\\
	&= \Delta(X^*_t, \bar{\theta}_t) + \Delta(X_t, \bar{\theta}_t) + \langle \bar{\theta}_t, X^*_t - X_t\rangle\,,
\end{align*}
where we define for $x\in\cX$ and $\theta\in\Theta$, $\Delta(x, \theta) = \lvert \langle \theta^*-\theta, x\rangle\rvert$ the absolute error made by $\theta$ with respect to the true parameter of the linear bandit $\theta^*$ in the direction of $x$. If we know a sequence of functions $\gamma_t\colon \cX\rightarrow \R_+$ such that with high probability, for all $t\in\N$ and $x\in\cX_t$, $\Delta(x, \bar{\theta}_t) \leq \gamma_t(x)$, then the principle of optimism recommends the action $X_t\in\argmax_{x\in\cX_t} \langle \bar{\theta}_t, x\rangle + \gamma_t(x)$, i.e., the one leading to the best plausible reward with respect to the confidence on the prediction error of $\bar{\theta}_t$. In this case, $r_t\leq \Delta(X^*_t, \bar{\theta}_t) + \Delta(X_t, \bar{\theta}_t) + \gamma_t(X_t) - \gamma(X^*_t)\leq 2\gamma_t(X_t)$ with high probability, and hence $\cR_T \leq 2\sum_{t=1}^T \gamma_t(X_t)$. We detail below how Corollary~\ref{corr:conf} coupled with standard assumptions provides such a bound.

\paragraph{Bound on the Prediction Error}

We follow the standard strategy of decoupling the dependency on $\bar{\theta}_t$ and $x$ in $\Delta(x, \bar{\theta}_t)$. By Cauchy-Schwarz's inequality, we have, for some positive definite matrix $P$ to be determined later,
\begin{align*}
\Delta(x, \bar{\theta}_t) = \lvert \langle P^{\frac{1}{2}}(\theta^*-\bar{\theta}_t), P^{-\frac{1}{2}}x\rangle\rvert \leq \lVert \theta^* - \bar{\theta}_t\rVert_{P} \lVert x \rVert_{P^{-1}}\,.
\end{align*}
As we see below, a natural choice for $P$ is the (average) Hessian of the empirical risk minimization problem, and therefore the term $\lVert x \rVert_{P^{-1}}$ can be handled by the elliptical potential lemma (Lemma 11 in \citet{abbasi_yadkori2011}). To control the remainder term in $\theta^* - \bar{\theta}_t$, we borrow technical tools from the classical approach developed for generalized linear bandits \citep{filippi2010parametric, faury2020improved} and note that 
\[F^{\alpha}_t(\theta^*) - F^{\alpha}_t(\bar{\theta}_t) = \bar{H}^{\alpha}_t(\theta^*, \bar{\theta}_t)(\theta^* - \bar{\theta}_t)\,,\]
where $\bar{H}^{\alpha}_t(\theta^*, \bar{\theta}_t) = \int_{0}^1 H^{\alpha}_t (u\theta^* + (1-u)\bar{\theta}_t)du$ is the average of the Hessian matrices along the segment $[\bar{\theta}_t, \theta^*]$\footnote{One could also use $\bar{H}^{\alpha}_t(\theta^*, \bar{\theta}_t) = \int_{0}^1 H^{\alpha}_t (\gamma_u)du$ where $\gamma\colon [0, 1] \rightarrow \Theta$ is smooth, unit speed path connecting $\bar{\theta}_t$ and $\theta^*$.} (this follows from the observation that the differential of $F^{\alpha}_t$ is  $H^{\alpha}_t$). Therefore the choice $P=\bar{H}^{\alpha}_t(\theta^*, \bar{\theta}_t)$ yields 
\begin{align*}
\lVert \theta^* - \bar{\theta}_t\rVert_{P} &= \lVert F^{\alpha}_t(\theta^*) - F^{\alpha}_t(\bar{\theta}_t)\rVert_{\bar{H}^{\alpha}_t(\theta^*, \bar{\theta}_t)^{-1}} \\
&\leq \lVert F^{\alpha}_t(\theta^*) - F^{\alpha}_t(\wh{\theta}_t)\rVert_{\bar{H}^{\alpha}_t(\theta^*, \bar{\theta}_t)^{-1}} \\
&\quad + \lVert  F^{\alpha}_t(\bar{\theta}_t) - F^{\alpha}_t(\wh{\theta})\rVert_{\bar{H}^{\alpha}_t(\theta^*, \bar{\theta}_t)^{-1}}\,.
\end{align*}
To conclude, we need to find a way to relate the local metric defined by $\bar{H}^{\alpha}_t(\theta^*, \bar{\theta}_t)^{-1}$ to those defined by $H^{\beta}_t(\theta^*)^{-1}$ and $H^{\beta}_t(\bar{\theta}_t)^{-1}$, for which we have high confidence bounds. This motivates the following assumption.

\begin{restatable}[Transportation of Local Metrics]{lem}{lemmatransport}\label{lem:transport}
	Under Assumption~\ref{ass:loss_curvature}, for $\alpha\!>\!0$, there exists $\kappa\!>\!0, \beta\!>\!0$ such that
	\[\bar{H}^{\alpha}_t(\theta^*, \bar{\theta}_t) \succcurlyeq \frac{1}{\kappa} H^{\beta}_t(\theta^*) \quad \text{and}\quad \bar{H}^{\alpha}_t(\theta^*, \bar{\theta}_t) \succcurlyeq \frac{1}{\kappa} H^{\beta}_t(\bar{\theta}_t)\,.\]
\end{restatable}

We detail in Appendix~\ref{app:C} that a suitable choice of parameter is $\beta=\kappa\alpha$ with $\kappa=\frac{M}{m}$ the conditioning of the loss $\cL$, which is a direct consequence of Assumption~\ref{ass:loss_curvature}. Again, we keep the formulation fairly generic as Lemma~\ref{lem:transport} may hold beyond losses with bounded curvature. For instance, in a special case of GLB, namely the logistic bandit, it is shown in \citet{faury2020improved} that this lemma holds thanks to self-concordance properties of the sigmoid link function. 

\subsection{Examples}\label{sec:examples_loss}

We conclude this 	section by discussing examples of standard losses and whether they satisfy the above conditions.

\begin{example}[Expectile]
	The expectile loss is derived from the potential $\psi_2(z)=\lvert p - \ind_{z<0}\rvert z^2$, the second derivative of which is $\psi_2''(z)=2\lvert p - \ind_{z<0}\rvert$. Thus, Assumption~\ref{ass:loss_curvature} holds with $m=2\min(p, 1-p)$ and $M=2\max(p, 1-p)$.
\end{example}

\begin{example}[Quantiles]
	The quantile loss is derived from the potential $\psi_1(z)=(p - \ind_{z<0}) z$, which is piecewise linear. In particular, it is not strongly convex and thus does not satisfy Assumption~\ref{ass:loss_curvature}. Bandits with quantile regression are therefore outside the scope of this work.
\end{example}

\subsection{Regret Analysis}

We make two additional standard assumptions that prior bounds are known on $\theta^*$ and on the actions $\cX=\bigcup_{t\in\bN} \cX_t$, which is standard in the existing literature on linear bandits.

\begin{restatable}[Prior Bound on Parameters]{ass}{assumptionfour}\label{ass:4} 
	All parameters are bounded by $S$, i.e., $\Theta \subseteq \cB^d_{\lVert \cdot \rVert_2}(0, S)$. In particular, this implies that $\lVert \theta^*\rVert_{H^{\beta}_t(\theta^*)^{-1}}\leq \frac{S}{\sqrt{\beta}}$ for any $\beta>0$.
\end{restatable}

\vspace*{-3mm}

\begin{restatable}[Prior Bound on Actions]{ass}{assumptionfive}\label{ass:5} 
	All actions are bounded by $L$, i.e., $\cX \subseteq \cB^d_{\lVert \cdot \rVert_2}(0, L)$.
\end{restatable}

We now obtain a high probability upper bound on the regret incurred by Algorithm~\ref{algo::UCB_convex_risk} for an explicit choice of exploration bonus sequence $(\gamma_t)_{t\in\N}$ and projection $\Pi$. As is standard for contextual bandits, this bound is \textit{minimax} (worst-case) as it does not depend explicitly on the optimality gaps $\langle \theta^*, X^*_t\rangle - \langle \theta^*, X_t\rangle$.


\begin{restatable}[Regret upper bound for LinUCB-CR - 1]{thm}{theoremone}\label{thm:regret_ucb}
	Let $\delta\!\in\!(0, 1)$, $\alpha\!\geq\!\max(1, L^2)$ and define for $t\!\in\!\N$ the exploration bonus
	\begin{align*}
	&\gamma_t\colon x\in\cX_t \mapsto c^\delta_t \lVert x \rVert_{H^{\kappa \alpha}_t(\bar{\theta}_t)^{-1}}\,,\\
	&c^\delta_t=2\kappa\left(\sigma\sqrt{2\log\frac{1}{\delta} + d\log\frac{m}{\alpha} + \log\det V^{\frac{\alpha}{m}}_t} + \sqrt{\frac{\alpha}{\kappa}}S\right)
	\end{align*}
	and the projection operator 
	\begin{align*}
	\Pi\colon \wh{\theta}\in\R^d \mapsto \argmin_{\theta\in\Theta} \lVert F^{\alpha}_t(\theta) - F^{\alpha}_t(\wh{\theta})\rVert_{H^{\kappa \alpha}_t(\theta)^{-1}}\,.
	\end{align*}
	
	Under Assumptions~\ref{ass:loss_curvature}-\ref{ass:subgauss}-\ref{ass:4}-\ref{ass:5}, with probability at least $1-\delta$, the regret of Algorithm~\ref{algo::UCB_convex_risk} is bounded by
\[\cR_T \leq 2c^\delta_T \max\left(\frac{1}{\sqrt{m}}, \frac{L}{\sqrt{\kappa \alpha}}\right) \sqrt{2Td\log\left(\!1\!+\!\frac{m T L ^2}{d\kappa\alpha}\right)}\,. \]
	In particular, we have $\cR_T = \cO\left(\frac{\kappa\sigma d}{\sqrt{m}}\sqrt{T}\log \frac{TL^2}{d}\right)$.
\end{restatable}

The proof of this result follows the standard regret analysis of LinUCB, up to the modification detailled in the previous sections. We report the detailed arguments in Appendix~\ref{app:proof_linucb}.

\begin{remark}
	This regret bound scales with $\kappa m^{-1/2}$, where $m$ is the minimum curvature of the loss $\cL$ and $\kappa$ the coefficient of transportation of local metrics. Under Assumption~\ref{ass:loss_curvature}, this scales as $m^{-3/2}$. In the limit of flattening loss $m\!\rightarrow\!0$, learning with this strategy becomes impossible. We show in Appendix~\ref{app:proof_linucb} that a small modification of the exploration sequence reduces this dependency to $\kappa^{1/2}m^{-1/2}$, at the cost of loosing local information carried by $H^{\kappa \alpha}_t(\bar{\theta_t})$. An analogous dependency on $m^{-1}$ was observed for GLB in \citet{filippi2010parametric}. In the special case of logistic bandit, \citet{faury2020improved} obtained a $\kappa$ independent of $m$ using self-concordance, and even pushed the dependency on $m^{-1/2}$ to higher order terms in $T$ using a more intricate algorithmic design. We conjecture that a similar construction could apply here but leave this open for future work.
\end{remark}

\begin{remark}
	In the mean-linear case, $m\!=\!M\!=\!\kappa\!=\!1$ and $H^\alpha_t\!=\!V^\alpha_t$, thus this result is compatible with the minimax lower bound $\cO(d\sqrt{T})$ for actions in $\cB^d_{\lVert \cdot \rVert}(0, 1)$ \citep[Theorem~24.2]{lattimore2020bandit} and matches the standard LinUCB upper bound \citep{abbasi_yadkori2011}.
\end{remark}

Theorem~\ref{thm:regret_ucb} holds for arbitrary (potentially adversarial) sequence of action sets $\left(\cX_t\right)_{t\in\bN}$. If these are instead stochastically generated, the regret bound can be further tightened.

\begin{restatable}[Stochastic action sets]{ass}{assumptionsix}\label{ass:6} Let $\nu_{\cX}$ a probability measure on $2^{\cB^d_{\lVert \cdot \rVert_2}(0, L)}$ (i.e., samples drawn from $\nu_{\cX}$ are sets of vectors of $L^2$ norm at most $L$). 
\\
\hspace*{2mm}
\textit{(i)} For $t\!\in\!\N$, $\cX_t\!\sim\!\nu_{\cX}$ defines an i.i.d. sequence of random action sets.
\\
\hspace*{2mm}
\textit{(ii)} Recall that $X_t\!\in\!\cX_t$ denotes the action selected by the agent at time $t\in\bN$. Then $\bE\left[X_{t} X_{t}^\top \lvert \cF_{t-1} \right]\!\succcurlyeq\!\rho_{\cX}L^2I_d\!>\!0$.
\end{restatable}

\begin{restatable}[Regret upper bound for LinUCB-CR - 2]{thm}{theoremonebis}\label{thm:regret_ucb_bis}
	Under Assumptions~\ref{ass:loss_curvature}-\ref{ass:subgauss}-\ref{ass:4}-\ref{ass:5}-\ref{ass:6}, with probability at least $1-2\delta$, the regret of Algorithm~\ref{algo::UCB_convex_risk} is bounded by
	\[\cR_T \leq 4 c^\delta_T\sqrt{\frac{2T}{m \rho_{\cX}}}\left(1 +\frac{C}{\sqrt{T}}\right)\,, \]
	where $C$ is a constant independent of $T$.
	In particular, we have $\cR_T = \cO\left(\kappa\sigma \sqrt{\frac{ dT}{m\rho_{\cX}}\log \frac{TL^2}{d}}\right)$.
\end{restatable}

The lower bound on conditional covariance of actions of Assumption~\ref{ass:6} is new, although related to more standard settings. In the case of finite action sets $\cX_t=\left\lbrace X_{k, t}, k\in[K]\right\rbrace$, \citet{li2017provably, kim2022double} considered a lower bound on the unconditional average covariance across arms $\bE\left[ \frac{1}{K}\sum_{k\in[K]} X_{k, t} X_{k, t}^\top\right]$. We argue that this assumption is quite mild in the sense that for non-degenerate $\nu_{\cX}$, the conditioning is essentially irrelevant. At time $t$, $\mathcal{X}_t$ is drawn independently of $\mathcal{F}_{t-1}$, and $X_t\in\mathcal{X}_t$ is selected in a $\mathcal{F}_{t-1}$-measurable fashion. To violate the covariance inequality, there should exist a fixed strict subspace $\mathcal{V}\subset \mathbb{R}^d$ such that with some probability $\mathcal{X}_t \cap \mathcal{V} \neq \emptyset$ (when randomizing over the action set $\mathcal{X}_t$) and $X_t$ should be one of the vectors in $\mathcal{V}$; however, if, e.g., $\nu_{\cX}$ spans an open set, this almost surely cannot happen. In other words, Theorem~\ref{thm:regret_ucb_bis} shows that if action sets $\cX_t$ are generated with enough diversity and no adversarial bias, the regret of optimistic strategies can be slightly improved by a factor $\cO(\log(T))$. Finally, note that Assumption~\ref{ass:6} and $\rho_{\mathcal{X}}$ do not influence the design of Algorithm~1, only the $\cO\left(\log T\right)$ term in its regret upper bound. 

In general, $\rho_{\cX}^{-1}\geq d$ and in many cases $\rho_{\cX}^{-1}=\cO(d)$ (see Appendix~\ref{app:proof_ucb_bis}), hence the regret upper bound scales linearly with $d$. Compared to \cite{kim2022double}, our proof relies on line crossing arguments developed in \cite{howard2020time} rather than on a crude union bound, leading to improved constants and higher order terms (even in the mean-linear case). We refer to Appendix~\ref{app:proof_ucb_bis} for additional details.

%

%
\section{APPROXIMATE STRATEGY WITH ONLINE GRADIENT DESCENT}
So far, we have shown that the standard LinUCB principle can be extended to the convex loss setting with similar regret guarantees under some curvature assumption. However, this comes at the cost of a significant computational overhead since the estimator $\wh{\theta}_t$ needs to be calculated from scratch at each step as $\argmin_{\theta\in \R^d} \sum_{s=1}^{t-1} \cL(Y_s, \langle \theta, X_s\rangle) + \frac{\alpha}{2}\lVert \theta \rVert_2^2$. As a reminder, in the standard mean-linear case, this estimator has an analytical expression that amounts to incrementally inverting the matrix $V^\alpha_t$, which can be done efficiently from the knowledge of the inverse of $V^{\alpha}_{t-1}$ via the Sherman-Morrison formula.

We propose an alternative algorithm that exploits online gradient descent (OGD) to compute a fast approximation of the empirical risk minimizer $\wh{\theta}_t$. This may be of practical interest to deploy risk-aware linear bandits in time-sensitive environments, such as in real-time online recommendation systems. Moreover, it can also be relevant in the mean-linear setting with high dimensional action sets, where computing gradients may be more tractable than inverting a large $d\times d$ matrix. For $\wh{\theta}\in\bR^d$, we use the shorthand $\nabla^{\alpha}_{n,h}(\wh{\theta})=\sum_{k=1}^{h} \partial\cL(Y_{(n-1)h+k}, \langle \wh{\theta}, X_{(n-1)h+k}\rangle) \!+\! \alpha \wh{\theta}$.

\begin{algorithm}[]
	\caption{LinUCB-OGD-CR\label{algo::UCB-OGD_convex_risk}}
	\SetKwInput{KwData}{Input} 
	\KwData{horizon $T$, regularisation parameter $\alpha$, projection $\Pi$, exploration bonus sequence $(\gamma^{\text{OGD}}_{t, T})_{t\leq T}$, step sequence $(\epsilon_t)_{t\in\N}$, episode length $h>0$.}
	\SetKwInput{KwResult}{Initialization}
	\SetKwComment{Comment}{$\triangleright$\ }{}
	\SetKwComment{Titleblock}{// }{}
	\KwResult{Observe $\cX_1$, set $\wh{\theta}^{\text{OGD}}_0$, $t=1$, $n=1$.} 
	\For{$t=1, \dots, T$}{
		\If{$t=nh+1$}{
				$\widehat{\theta}^{\text{OGD}}_{n}= \widehat{\theta}^{\text{OGD}}_{n-1} - \epsilon_{n-1} \nabla^{\alpha}_{n,h}(\wh{\theta}^{\text{OGD}}_{n-1})$ \Comment*[r]{OGD} 
			$\bar{\theta}^{\text{OGD}}_{n} =  \frac{1}{n}\sum_{j=1}^n\Pi(\wh{\theta}^{\text{OGD}}_j)$ \Comment*[r]{Average}
			$n \leftarrow n+1$\,
		}	
a		$X_t = \arg\max_{x\in\cX_t} \langle\bar{\theta}^{\text{OGD}}_{n}, x\rangle + \gamma^{\text{OGD}}_{t, T}(x)$ \Comment*[r]{Play with same parameter for $h$ steps}
		Observe $Y_t$ and $\cX_{t+1}$\,;\\
		$t \leftarrow t + 1$\,;
	}
\end{algorithm}

The intuition behind Algorithm~\ref{algo::UCB-OGD_convex_risk} is that at time $t=nh\!+\!1$, the approximation error between the OGD estimate $\bar{\theta}^{\text{OGD}}_n$ and the exact minimizer of the empirical risk $\widehat{\theta}_t$ induces additional exploration, which translates into an increased regret compared to LinUCB. In other words, LinUCB-OGD trades off accuracy for computational efficiency. The episodic structure is borrowed from \cite{ding2021efficient} and is key to ensure sufficient convexity of the aggregate loss $\nabla^{\alpha}_{n,h}(\wh{\theta})$. This allows to leverage the strong approximation guarantees of OGD, which we extend in the following proposition by relaxing the standard boundedness requirement of the gradient (Theorem~3.3, \citet{hazan2019introduction}) to a weaker sub-Gaussian control at a given parameter. We prove in Appendix \ref{appendix:proof_sgd} an extension of the following proposition, with an explicit bound on the OGD regret, which we report below in the $\cO$ notation for the sake of concision.

\begin{restatable}[OGD Regret, Sub-Gaussian Gradients]{prop}{propogd}\label{prop:ogd}
	Let $\cC$ a convex subset of $\R^d$ and $\Pi$ the projection operator onto $\cC$. For $j=1, \dots, N$, let $\ell_j\colon \cC \longrightarrow \R_+$ a twice differentiable convex function and $a, A >0$ such that $a I_d \preccurlyeq \nabla^2 \ell_j(z) \preccurlyeq A I_d$ for all $z\in\cC$. Define the OGD update at step $j$ by $z_j=\Pi(z_{j-1} - \epsilon_{j-1}\nabla \ell_j(z_{j-1}))$ and $\bar{z}_n=\arg\min_{z\in\cC} \sum_{j=1}^n \ell_j(z)$. Assume that there exists $z^*\in\cC$ such that $\nabla \ell_j(z^*) = g_j + \frac{\alpha}{n} z^*$ with $\alpha\geq 0$ and $g$ a centered, $\R^d$-valued $\sigma$-sub-Gaussian process, and also that $\cC$ is bounded, i.e., $\text{diam}(\cC) = \sup_{z,z'\in\cC} \lVert z-z'\rVert<\infty$.
	Then with probability at least $1\!-\!\delta$, the OGD regret with step size $\epsilon_j\!=\!\frac{3}{aj}$ is bounded uniformly in $N\rightarrow +\infty$ by 
	\[\sum_{j=1}^N \ell_j(z_j) - \ell_j(\bar{z}_N) = \cO\left(\frac{d\sigma^2}{a}\log^2 N\right)\,.\]
	
\end{restatable}

Our final result, which we prove in Appendix~\ref{appendix:proof_ucb-ogd}, states that the approximation error of OGD induces at most a polylog correction in the regret of LinUCB-OGD-CR.

\begin{restatable}[Regret of LinUCB-OGD-CR]{thm}{theoremtwo}\label{thm:regret_ucb-ogd}
	Let $\epsilon_h>0$ and $h=\lceil \frac{2\epsilon_h}{\rho_{\cX}L^2} + \frac{8}{\rho_{\cX}^2}\log\frac{2}{\delta}\rceil$. Assume that $\partial \cL(Y_t, \langle\theta^*, X_t\rangle)$ is $\sqrt{m}\sigma$-sub-Gaussian for all $t\leq T$.
	Under Assumptions~\ref{ass:loss_curvature}-\ref{ass:subgauss}-\ref{ass:4}-\ref{ass:5}-\ref{ass:6}, there exists constants $C, C'>0$ such that with probability at least $1-(1+T/h)\delta$ the regret of Algorithm~\ref{algo::UCB-OGD_convex_risk} with exploration bonus sequence
	\begin{align*}
 	&\gamma^{\text{OGD}}_{t, T}\colon x\in\cX_t \mapsto (c^{\delta}_t + c^{\text{OGD}, \delta}_{t, T}) \lVert x \rVert_{H^{\kappa \alpha}_{t}(\bar{\theta}^{\text{OGD}}_{\lfloor\!\frac{t\!-\!1}{h} \!\rfloor})^{-1}}\,,\\
 	&c^{\text{OGD}, \delta}_{t, T}\!=\!\sqrt{\!\left(L^2\!+\!\frac{\alpha}{mMt}\right)\!\!\left(\!\frac{2\kappa C'd h^2\sigma^2}{\epsilon_h^2}\!\log\left(\!\frac{2dT}{h\delta}\!\right)\!\log\!\left(\!\frac{t}{h}\!\right)\!\right)\!}\,,
	\end{align*} 
	and the $OGD$ step sequence of Proposition~\ref{prop:ogd} satisfies
	\[\cR_T = \cO\left(\sigma \sqrt{\frac{\kappa dT}{m\rho_{\cX}}}\left(\sqrt{\kappa \log\left(\frac{TL^2}{d}\right)} + h \log\left(dT\right)\right)\right)\,.\]
\end{restatable}


The episode length $h$ scales as $\cO(\rho_{\cX}^{-2})$, which grows at least as fast as $\cO(d^2)$ in the action dimension $d$. This is sufficient to bound with high probability the smallest eigenvalue of the Hessian of the aggregate losses $\nabla^{\alpha}_{n,h}$ and thus ensure their strong convexity. However, longer episodes also means less frequent updates of $\bar{\theta}^{\text{OGD}}_n$, i.e., less learning, which is materialized by the additional dependency on $h$ in the regret. In Appendix~\ref{appendix:proof_ucb-ogd}, Lemma~\ref{lem:smallest_eig_tighter}, we deduce a tighter, more intricate expression for $h$, although still scaling as $\cO(\rho_{\cX}^{-2})$. We only report the simpler expression here to avoid cluttering.

\begin{remark}
	The union bound used in Proposition~\ref{prop:ogd} imposes the knowledge of the horizon $T$ at runtime (in the definition of $\gamma_{t,T}$), thus making Algorithm~\ref{algo::UCB-OGD_convex_risk} not anytime.
\end{remark}

%
%
%
%
%
%
%
%

\section{EXPERIMENTS}\label{sec:xp}

\begin{figure}
	\centering
	\begin{subfigure}[b]{0.48\textwidth}
		\centering
		\includegraphics[width=0.99\textwidth]{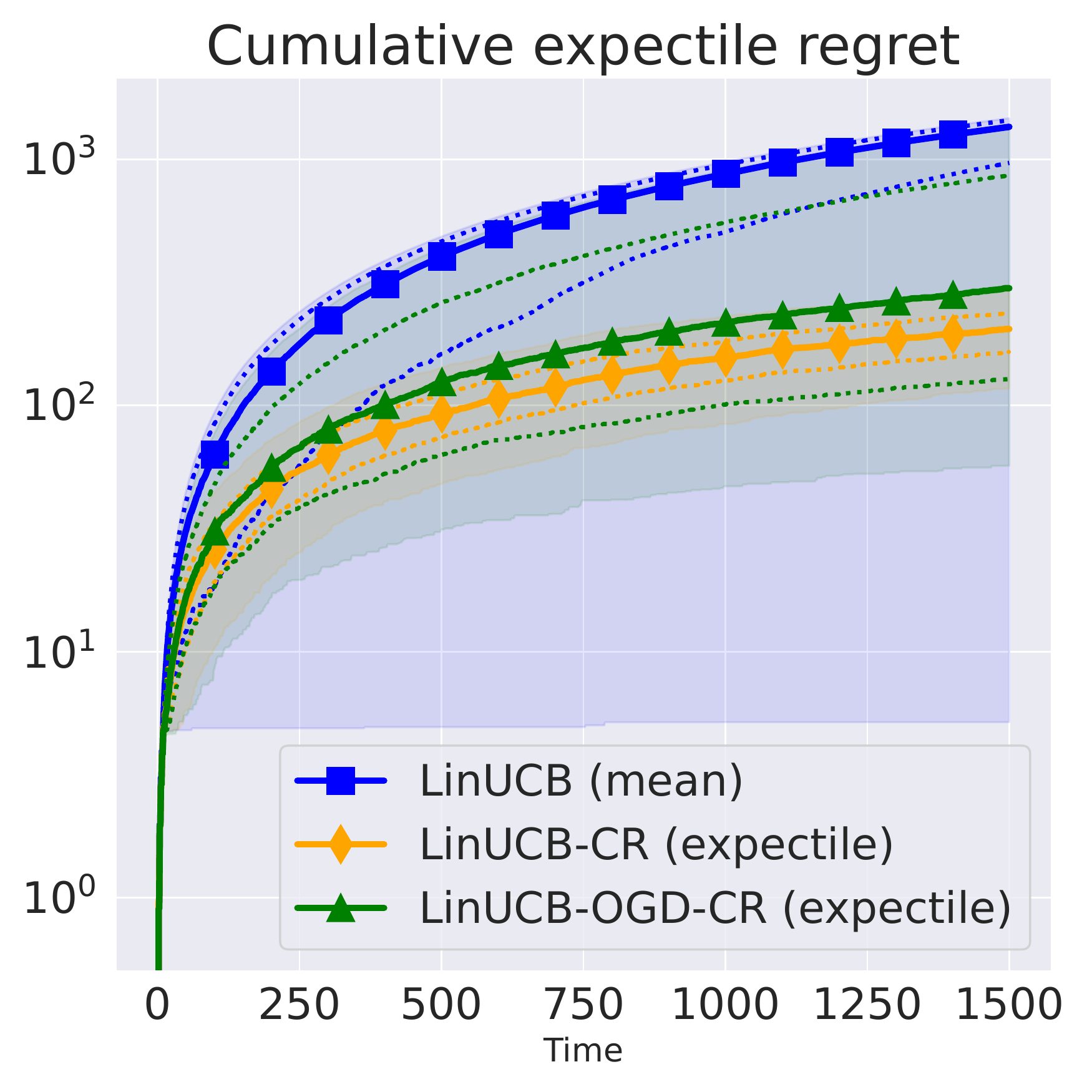}
	\end{subfigure}
	\caption{Two-armed linear $10\%$-expectile bandit with $\bR^3$ contexts and expectile-based asymmetric noises. Thick lines denote median cumulative regret over 500 independent replications. Dotted lines denote the 25 and 75 regret percentiles. Shaded areas denote the 5 and 95 percentiles.}
	\label{fig:xp_expectile}
\end{figure}

We conducted three numerical experiments to illustrate the performance of both risk-aware algorithms, under expectiles and entropic risk criteria. In Figure~\ref{fig:xp_expectile}, we considered an expectile-based asymmetric distribution \citep{torossian2020bayesian} with context-dependent $p$-expectiles. This distribution is log-concave, thus fitting the scope of the supermartingale control of Lemma~\ref{lem:supermart}. As recalled in Section~\ref{sec:examples_loss}, the corresponding loss satisfies Assumption~\ref{ass:loss_curvature} with $m=2\min(p, 1\!-\!p)$ and $M=2\max(1\!-\!p, p)$. Note that the more risk-averse ($p\rightarrow 0$), the flatter the loss ($m\rightarrow 0$) and thus the harder it is to learn. This matches the intuition on risk-aware measures: by focusing on the more extreme events, they require more samples to reach the same statistical accuracy. More details, including the other two experiments, are postponed to Appendix~\ref{app:xp}. As far as we are aware, no algorithm exists for the expectile criterion; for entropic risk, \citet{maillard2013robust} analyzes a variant of KL-UCB but only for the non-contextual multi-armed bandit problem (and without numerical evidences).

The settings of these numerical experiments were designed so that the optimal arms were different depending on the criterion of interest (mean versus risk-aware). Instances of the classical LinUCB algorithm \citep{abbasi_yadkori2011} were indeed deceived and accumulated linear risk-aware regret, while Algorithms~\ref{algo::UCB_convex_risk} and \ref{algo::UCB-OGD_convex_risk} exhibited milder sublinear trends. As expected, the LinUCB-OGD variant accumulated slightly more regret and showed higher variability across independent replications compared to LinUCB with the exact minimization of the empirical risk, at the benefit of improved runtimes (Table~\ref{table:runtime_linear_expectile}).


\begin{small}
	\begin{table}[!ht]
		\caption{Runtimes for the Classical LinUCB and Algorithms~\ref{algo::UCB_convex_risk} (LinUCB for Convex Risk) and \ref{algo::UCB-OGD_convex_risk} (LinUCB-OGD for Convex Risk), Reported in Seconds as Mean $\pm$ Standard Deviation, Estimated Across 500 Independent Replications with Time Horizon $T=1500$.}
		\label{table:runtime_linear_expectile}
		\centering
		\begin{tabular}{llll}
			\toprule
			Algorithm    & Runtime \\
			\midrule
			LinUCB (mean) & $37.2 \pm 4.9$ \\[1.0ex]
			LinUCB-CR (expectile) & $814.8 \pm 88.3$ \\[1.0ex]
			LinUCB-OGD-CR (expectile) & $60.2 \pm 12.0$ \\[1.0ex]
			\bottomrule
		\end{tabular}
		\vskip -5mm
	\end{table}
\end{small}

\section{CONCLUSION}
We have introduced a new setting for contextual bandits, building on the recent interest for risk-awareness in multi-armed bandits. We reviewed the literature on risk measures, in particular the notion of elicitability, that allows to extend the risk minimization framework of ridge regression beyond standard mean-linear bandits. To lift the regret analysis of optimistic algorithms to the setting of scalar risk measures $\rho_{\cL}$ elicited by a convex loss $\cL$, we showed that uniformly bounding the curvature of the loss (Assumption~\ref{ass:loss_curvature}) is sufficient to maintain satisfying theoretical guarantees ($\cO(\sqrt{T})$ worst-case regret, up to polylog terms (Theorem~\ref{thm:regret_ucb} and \ref{thm:regret_ucb_bis}). More precisely, we identified two key conditions, namely a supermartingale control (Lemma~\ref{lem:supermart}) and a transportation inequality (Lemma~\ref{lem:transport}), that guarantee sublinear regret; while these are direct consequences of the bounded curvature assumption, they may hold in different settings, as was recently discovered in GLB. 

Going further, we believe it would be interesting to extend the linear model between actions and risk measures to generalized linear models ($\rho_{\cL}(Y_t) = \mu(\langle \theta, X_t\rangle)$ for some link function $\mu\colon \R\rightarrow \R$), kernelized bandits ($\rho_{\cL}(Y_t) = f(X_t)$ where $f$ belongs to some RKHS) or neural bandits ($\rho_{\cL}(Y_t) = f_{\theta}(X_t)$ where $f_{\theta}$ is a neural network with weights $\theta$). Moreover, capturing well-established risk measures such as mean-variance, conditional value-at-risk or quantiles would require to adapt the theory to high-order elicitable measures and to non-smooth losses. Finally, we believe the technical results developed here, in particular the supermartingale control and the transportation inequality, can pave a way for the design and analysis of Thompson sampling strategies in the contextual risk-aware setting.

\subsubsection*{Acknowledgements}
The authors acknowledge the funding of the French National Research Agency, the French Ministry of Higher Education and Research, Inria, the MEL and the I-Site ULNE regarding project R-PILOTE-19-004-APPRENF and Bandits
For Health (B4H). We thank the anonymous reviewers for their careful reading of the paper and their suggestions
for improvements. Experiments presented in this paper were carried out using the Grid’5000 testbed, supported by
a scientific interest group hosted by Inria and including CNRS, RENATER and several universities as well as other
organizations (see https://www.grid5000.fr). We also thank the organizers and attendees of the European Workshop on Reinforcement Learning (EWRL) 2022, where this work was initially presented.

\bibliographystyle{abbrvnat}
\bibliography{references}

\newpage
\onecolumn

\newpage

\appendix

\section{SUMMARY AND INTERPRETATION OF ELICITABLE RISK MEASURES}\label{app:elicitable}

We report in Table~\ref{table:elicitable} an overview of common elicitable risk measures and their associated loss functions. We recall that for a distribution $\nu$ over $\R$ and a loss function $\cL\colon \R\times \R^p\rightarrow \R$, we defined the risk measure elicited by $\cL$ as $\rho_{\cL}(\nu) = \arg\min_{\xi\in \R^p}\bE_{Y\sim\nu}\left[ \cL(Y, \xi)\right]$. Note that the pairs (mean, variance) and (VaR, CVaR) are second-order elicitable but neither the variance nor the CVaR are first-order elicitable. For these pairs, we report the generic form of elicitation losses, which depend on arbitrary convex functions $\psi_1$ and $\psi_2$, as well as instances of such losses obtained for the natural choice $\psi_1(\xi)=\psi_2(\xi)=\xi^2/2$.

\begin{small}
	\begin{table}
		\caption{Example of Elicitable Risk Measures.}
		\label{table:elicitable}
		\centering
		\begin{tabular}{llll}
			\toprule
			Name     & $\rho_{\cL}(\nu)$     & Associated loss $\cL(y, \xi)$ & Domain \\
			\midrule
			Mean & $\bE_{Y\sim\nu}[Y]$ & \makecell[l]{$(y-\xi)^2$\\ \\Bregman divergence $\cB_{\psi}(y, \xi)$\\$\psi(y) - \psi(\xi)-\psi'(\xi)(y-\xi),$\\$\psi$ differentiable,\\strictly convex.\\ \\ \\} & $\xi\in \R$\\[5.0ex]
			Derived from potential $\psi$& $\argmin_{\xi\in\R}\bE_{Y\sim\nu}[\psi(Y-\xi)]$ & $\psi(y-\xi)$ & $\xi\in dom(\psi)$\\[5.0ex]
			\makecell[l]{Generalized moment\\$T\colon \R\rightarrow \R$} & $\bE_{Y\sim\nu}[T(Y)]$ & $\frac{1}{2}\xi^2 - \xi T(y)$ & $\xi\in\R$\\[5.0ex]
			\makecell[l]{Entropic risk, $\gamma\neq 0$\\ (Example 1,\\ \citet{embrechts2021bayes})\\ \\ }& $\frac{1}{\gamma}\log\bE_{Y\sim\nu}[e^{\gamma Y}]$ & $\xi + \frac{1}{\gamma}(e^{\gamma(y-\xi)}-1)$ & $\xi\in\R$\\[5.0ex]
			\makecell[l]{(mean, variance)\\ (Example 1.23,\\ \citet{brehmer2017elicitability})}& \makecell[l]{$\mu=\bE_{Y\sim\nu}[Y]$ \\$\sigma^2=\bE_{Y\sim\nu}[Y^2] - \mu^2$} & \makecell[l]{$\frac{1}{2}\xi_1^2 + \frac{1}{2}(\xi_2 + \xi_1^2)^2$\\$- \xi_1 y - (\xi_2 + \xi_1^2)y^2$\\ \\$-\psi_1(\xi_1)-\psi'_1(\xi_1)(y-\xi_1)$\\$- \psi_2(\xi_2+\xi_1^2)$\\$- \psi'_2(\xi_2+\xi_1^2)(y^2-\xi_2-\xi_1^2))$,\\ $\psi_1, \psi_2$ differentiable,\\strictly convex.\\ \\ \\ \\ } & \makecell[l]{$\xi_1\in\R$\\ $\xi_2\geq 0$}\\[5.0ex]
			\makecell[l]{$(\text{VaR}_{\alpha}, \text{CVaR}_{\alpha})$,\\$\alpha\in(0, 1)$\\ (Corollary 5.5,\\ \citet{fissler2016higher})}& \makecell[l]{$\text{VaR}_{\alpha}=\inf\lbrace y\in\R, \int_{-\infty}^{y} d\nu \geq \alpha \rbrace$ \\$\text{CVaR}_{\alpha} = \frac{1}{\alpha} \int_{0}^{\alpha} \text{VaR}_a da$} & \makecell[l]{$(\xi_1-y)_+ - \alpha \xi_1$ \\ $+ \xi_2( \frac{1}{\alpha}(\xi_1-y)_+ - \xi_1 )$\\$+\frac{1}{2}\xi_2^2$\\ \\ $(\ind_{y\leq\xi_1} -\alpha) \psi'_1(\xi_1)$\\$- \ind_{y\leq \xi_1}\psi'_1(y)$ \\ $+ \psi'_2(\xi_2)(\xi_2\!-\!\xi_1\!+\!\frac{1}{\alpha}\ind_{y\leq\xi_1}(\xi_1\!-\!y))$\\$- \psi_2(\xi_2) + c(y)$,\\ $\psi_1$ convex,\\ $\psi_2$ strictly convex and increasing,\\ $c\colon \R \rightarrow \R$.\\ \\ } & $\xi_1\geq \xi_2$\\[5.0ex]
			\bottomrule
		\end{tabular}
		\vskip -5mm
	\end{table}
\end{small}

We provide below some intuition about these commonly used measures in risk management.

\paragraph{Mean-Variance}
Assessing the risk-reward tradeoff of an underlying distribution $\nu$ by penalizing its mean by a higher order moment (typically the variance) is perhaps the most intuitive of risk measures. Following \citet{markowitz1952march}, the mean-variance risk measure at risk aversion level $\lambda\in\R$ is defined by $\rho_{MV_1}(\nu) = \mu - \lambda\sigma$, where $\mu$ and $\sigma$ denote the mean and standard deviation of $\nu$. Alternatively, it can also be defined as $\rho_{MV_2}(\nu) = \mu - \frac{\lambda}{2}\sigma^2$, using the variance rather than the standard deviation in the penalization term. Both measures are especially well-suited for Gaussian distributions as $\mu$ and $\sigma$ fully characterize this family.

\paragraph{VaR and CVaR}
For a distribution with continuous cdf (i.e., it has no atom), the Value-at-Risk $\text{VaR}_{\alpha}(\nu)$ at level $\alpha\in(0, 1)$ is equivalent to the $\alpha$ quantile, and a simple change of variable reveals that the Conditional Value-at-Risk $\text{CVaR}_{\alpha}(\nu)$ is thus $\bE\left[X\mid X\leq \text{VaR}_\alpha(\nu)\right]$. Intuitively, a random variable with a high $\text{CVaR}_{\alpha}$ distribution takes on average relatively high values in the "$\alpha\%$ worst-case" scenario. For $\alpha\rightarrow 1^-$, $\text{CVaR}_{\alpha}(\nu) \rightarrow \bE_{Y\sim\nu}[Y]$ and thus the risk measure becomes oblivious to the tail risk; on the contrary, the case $\alpha\rightarrow 0^+$ emphasizes only the worst outcomes. 

In the Gaussian case $\nu\sim\cN(\mu, \sigma)$, using the notations $\phi$ and $\Phi$ respectively for the pdf and cdf of the standard normal distribution, simple calculus shows that
\begin{align*}
	\text{VaR}_{\alpha}(\nu) &= \mu + \sigma\Phi^{-1}(\alpha)\,,\\
	\text{CVaR}_{\alpha}(\nu) &= \mu - \frac{\sigma}{\alpha\sqrt{2\pi}}\phi\left(\Phi^{-1}(\alpha)\right)\,,\\
\end{align*}
i.e., $\text{CVaR}_{\alpha}(\nu) = \rho_{MV_1}(\nu)$ with risk aversion level $\lambda=\frac{1}{\alpha \sqrt{2\pi}}\phi\left(\Phi^{-1}(\alpha)\right)$. In particular, increasing the variance $\sigma^2$ reduces $\text{CVaR}_{\alpha}(\nu)$, corresponding to the intuition of higher volatility risk.

\paragraph{Entropic Risk}

The non-elicitability of $\text{CVaR}_{\alpha}$ motivated the use of the entropic risk as an alternative measure. This measures rewrites as (see \citet{brandtner2018entropic})
\[\rho_{\gamma}(\nu) = \sup_{\nu' \text{ probability measure}} \left\lbrace \bE_{Y\sim\nu'}[Y] - \frac{1}{\gamma}\text{KL}(\nu' \lVert \nu)\right\rbrace\,.\]

The intuition here is similar to the mean-variance measure, i.e., penalizing the expected value by a measure of uncertainty, but differs by the use of the Kullback-Leibler divergence $\text{\KL}(\nu'\lVert \nu) = \bE_{Y\sim\nu'}[ \log\frac{d\nu'}{d\nu}]$ instead of the variance. The entropic risk measure can be interpreted as the largest expected value that a mispecified model $\nu'$ (in place of the true underlying distribution $\nu$) may have, where $\text{\KL}(\nu'\lVert \nu)$ controls the magnitude of the mispecification.

Again, in the Gaussian case, this measure reduces to $\rho_{\gamma}(\nu) = \mu + \frac{\gamma}{2}\sigma^2=\rho_{MV_2}(\nu)$ at risk aversion level $\lambda=-\gamma$.

\paragraph{Expectile}
Beyond their interpretation as generalized, smooth quantiles, expectiles can also be understood in light of the financial risk management literature.
Let $e_p(\nu)$ denote the $p$-expectile of $\nu$ for a given probability $p\in(0, 1)$. Then, simple calculus shows that
\[(1-p) \bE_{Y\sim\nu}[(e_p(\nu) - Y)_+] = p \bE_{Y\sim\nu}[(Y-e_p(\nu))_+]\,,\]
where $z_+ = \max(z, 0)$. If $\nu$ represents the distribution of a tradeable asset $Y$ at time $T$, then the $p$-expectile is the strike $K=e_p(\nu)$ such that call and put on $Y$ struck at $K$ at maturity $T$ are in proportion $\frac{1-p}{p}$ to each other, where we define the call and put prices (with zero time discounting) by respectively
\begin{align*}
C(\nu, K) = \bE_{Y\sim\nu}[(Y-K)_+]\,,\\
P(\nu, K) = \bE_{Y\sim\nu}[(K-Y)_+]\,.
\end{align*}
Similarly, \citet{keating2002universal} introduced the notion of Omega ratio as a risk-return performance measures. It is defined at level $K$ by
\[\Omega(K) = \frac{\int_{K}^{+\infty} \left(1 - F(y)\right) dy}{\int_{-\infty}^{K} F(y)dy}\,,\]
where $F$ is the cdf of $\nu$. This ratio can also be viewed as a call-put ratio, hence another definition of the $p$-expectile is via the implicit equation $\Omega(K)=\frac{1-p}{p}$ for $K=e_p(\nu)$. 

Contrary to the previous risk measures, it may not be clear from this definition alone that expectiles do encode a notion of aversion to risk. The next proposition shows that $p$-expectiles of many distributions, including normal and adjusted lognormal, are decreasing functions of their variances when $p<\frac{1}{2}$, thus penalizing more volatile distributions, making them suitable for risk management. We provide an elementary proof using the tools of the financial mathematics literature, where such risk measures were extensively studied.

\begin{restatable}{prop}{prop_expectile_vega}\label{prop:expectile_vega} Let $\cI\subseteq \bR^*_+$ an open set and $\lbrace \nu_\sigma, \sigma\in\cI\rbrace$ a family of probability distributions such that
	\begin{enumerate}[(i)]
		\item the expectation mapping $\sigma\in\cI\mapsto \bE_{Y\sim\nu_\sigma}\left[Y\right]$ is constant,
		\item both the call and put mappings $C(\cdot, K)\colon \sigma\in\cI\mapsto \bE_{Y\sim\nu_\sigma}\left[\left(Y-K\right)_+\right]$ and $P(\cdot, K)\colon \sigma\in\cI\mapsto \bE_{Y\sim\nu_\sigma}\left[\left(K-Y\right)_+\right]$ are differentiable and nondecreasing, for any $K\in\bR$.
	\end{enumerate}
	For $p\in(0, 1)$, we let $e_p(\sigma)=e_p(\nu_\sigma)$. Then $\sign \frac{d}{d\sigma}e_p(\sigma) = \sign(p-\frac{1}{2})$.
\end{restatable}

Before we proceed to the proof, let us note that two classical families of distributions satisfy these assumptions (see \citep[Theorem~8]{merton1973theory} for a a general result).
\begin{itemize}
	\item \textbf{Normal}: for $\mu_0\in\bR$, $\lbrace \nu_\sigma=\cN(\mu_0, \sigma^2),\ \sigma\in\bR_+^* \rbrace$, for which $\bE_{Y\sim\nu_\sigma}[Y]=\mu_0$.
	\item \textbf{Adjusted lognormal}: for $\mu_0\in\bR$, $\lbrace \nu_\sigma=\exp\left(\cN(\mu_0, \sigma^2) - \frac{\sigma^2}{2}\right),\ \sigma\in\bR_+^* \rbrace$, for which $\bE_{Y\sim\nu_\sigma}[Y]=e^{\mu_0}$.
\end{itemize}

In particular in the normal case, it follows from Lemma~\ref{lemma:add_psi} that $e_p(\nu) = \mu_0 + \sigma e_p(\cN\left(0, 1\right))= \rho_{MV_1}(\nu_\sigma)$ at risk aversion level $\lambda=-e_p(\cN\left(0, 1\right))$ (positive if $p<\frac{1}{2}$). 

\proof{
We first recall the call-put parity principle, which states that for any distribution $\nu_\sigma$ and strike $K\in\bR$, the following equality holds:
\[C(\sigma, K) - P(\sigma, K) = \bE_{Y\sim\nu_\sigma}[Y] - K\,,\]
where we write $C(\nu_\sigma, K)=C(\sigma, K)$ and $P(\nu_\sigma, K)=P(\sigma, K)$. 

Notice that the call-parity principle and the assumption that $\frac{d}{d\sigma}\bE_{Y\sim\nu_\sigma}[Y]=0$ implies that $\partial_{\sigma}C = \partial_{\sigma}P$. We denote this quantity by $V$. First, note that $\sigma\in\cI \mapsto e_p(\sigma)$ is differentiable (implicit function theorem). From the equation $C(\sigma, e_p(\sigma)) = (1-p)/p P(\sigma, e_p(\sigma))$, we deduce that

\begin{align*}
\frac{d}{d\sigma} C(\sigma, e_p(\sigma)) &= \partial_{\sigma}C(\sigma, e_p(\sigma)) + \partial_K C(\sigma, e_p(\sigma)) \frac{d}{d\sigma}e_p(\sigma)\,,\\
\frac{d}{d\sigma} P(\sigma, e_p(\sigma)) &= \partial_{\sigma}P(\sigma, e_p(\sigma)) + \partial_K P(\sigma, e_p(\sigma)) \frac{d}{d\sigma}e_p(\sigma)\,,
\end{align*}
and thus
\begin{align*}
\frac{1-2p}{p} V + \frac{d}{d\sigma}e_p(\sigma) \left(\frac{1-p}{p}\partial_K P(\sigma, e_p(\sigma)) - \partial_K C(\sigma, e_p(\sigma))\right) = 0\,.
\end{align*}
Elementary option pricing principles show that $V\geq 0$, i.e., the call and put prices both increase with higher volatility, as well as $\partial_K C \leq 0$ and $\partial_K P \geq 0$. Therefore, we deduce that $\frac{d}{d\sigma}e_p(\sigma)\leq 0$.
$\hfill\blacksquare$
}

In particular for $p=1/2$, the $p$-expectile corresponds to the strike $K$ at which call and put have equal prices, which by the call-put parity principle (with zero discounting) implies that $K=\bE[Y]$, thus giving an alternative derivation of the equivalence between $1/2$-expectile and mean.

\section{PROPERTIES OF CONVEX LOSSES AND POTENTIALS}\label{app:convex_potentials}

Before we prove Lemma~\ref{lemma:linearity_psi}, we write the following technical lemma.

\begin{restatable}[Risk Measures $\rho_{\psi}$ Are Additive]{lem}{lemma_add_psi}\label{lemma:add_psi}
Let $\psi\colon \R\rightarrow \R$ be a strongly convex, differentiable function, $\nu$ be a distribution over $\R$ and $c\in\R$. Then $\rho_{\psi}(\nu+c) = \rho_{\psi}(\nu) + c$. 
\end{restatable}

\proof{
For the sake of simplicity, we assume $\nu$ admits a density $p$ (with respect to the Lebesgue measure) and that $\psi$ and $p$ are regular enough to allow for differentiation under the following integral. Then the risk measure associated with $\cL_{\psi}$ reads $\rho_{\psi}(\nu) = \argmin_{\xi\in\R} \int \psi\left(y - \xi \right)p(y) dy$ and the first order condition gives $\int \psi'(y - \rho_{\psi}(\nu))p(y)dy=0$. Similarly, for any $c\in\R$, we have $\int \psi'(y - \rho_{\psi}(\nu+c))p(y-c)dy=0$ since the density of $\nu+c$ is $p(\cdot - c)$. We now deduce from a simple change of variable $z=y-c$ that $\int \psi'(z + c - \rho_{\psi}(\nu+c))p(z)dz=0$, which shows that $\rho_{\psi}(\nu+c) - c$ is also a minimizer of $\xi\mapsto \int\psi\left(y - \xi \right)p(y) dy$. By uniqueness ($\psi$ is strongly convex), we deduce that $\rho_{\psi}(\nu+c) = \rho_{\psi}(\nu) + c$.
$\hfill\blacksquare$
}

\paragraph{Noise Additivity for Losses Derived From Potentials}

\begin{restatable}[]{lem}{lemma_lin_psi}\label{lemma:linearity_psi}
	Assume $\cL_{\psi}$ is adapted to the linear bandit $(\varphi, \theta^*)$ and $\psi$ is strongly convex and differentiable. Then there exists a stochastic process $\eta$ such that the bandit is represented at time $t$ by $Y_t\sim\langle \theta^*, X_t\rangle + \eta_t$ and $\rho_{\psi}(\eta\lvert \cF_t)=0$. 
\end{restatable}

\proof{
Define the process $\eta$ at time $t$ by $\eta_t=Y_t - \langle\theta^*, X_t\rangle$. To compute $\rho_{\psi}(\nu\lvert \cF_t)$, note that $X_t$ is measurable with respect to $\cF_t$, therefore by Lemma~\ref{lemma:add_psi} and the properties of conditional expectation, we have that $\rho_{\psi}(\eta_t\lvert \cF_t) = \rho_{\psi}\left(Y_t\lvert \cF_t\right)- \langle\theta^*, X_t\rangle=\rho_{\psi}\left(\varphi\left(\langle \theta^*, X_t \rangle\right)\lvert \cF_t\right)-\langle \theta^*, X_t \rangle=0$ by definition of $\cL_{\psi}$ being adapted to the bandit $(\varphi, \theta^*)$.

$\hfill\blacksquare$
}

\section{PROOF OF LEMMA~\ref{lem:supermart} AND PROPOSITION~\ref{prop:conf}}\label{app:proof_conf}

\lemsupermart*

\proof{Assumption~\ref{ass:loss_curvature} implies that $\partial^2 \cL^*_t \geq m$, therefore it is sufficient to show that there exists $\sigma>0$ such that
\[\bE\left[\exp\left(\langle\lambda, X_t\rangle \partial^1\cL^*_t - \frac{m\sigma^2}{2}\langle\lambda, X_t\rangle^2\right)\middle| \cF_t\right] \leq 1\,.\]
Since $X_t$ is $\cF_t$-measurable, this is equivalent to 
\[\bE\left[\exp\left(\langle\lambda, X_t\rangle \partial^1\cL^*_t \middle| \cF_t\right]\right) \leq \exp\left(\frac{m\sigma^2}{2}\langle\lambda, X_t\rangle^2\right) \,,\]
which follows from the sub-Gaussian property of the process $\partial^1\cL^*$ (Assumption~\ref{ass:subgauss}).

\proptwo*

\proof{The proof follows the method of mixture techniques, popularized in bandits by \cite{abbasi_yadkori2011}. For $\lambda\in\R^d$, we define the process $M^\lambda_t = \exp\left(\lambda^\top S_t -\frac{\sigma^2}{2}\lVert \lambda\rVert^2_{H^0_t(\theta^*)}\right)$. We recall the expression of the Hessian $H_t^0(\theta) = \sum\limits_{s=1}^{t-1}\partial^2 \cL\left(Y_s, \langle \theta, X_s\rangle\right)X_sX_s^\top$ and that in particular $\lVert \lambda\rVert^2_{H^0_t(\theta^*)} = \sum\limits_{s=1}^{t-1} \partial^2 \cL\left(Y_s, \langle \theta, X_s\rangle\right)\left(\lambda^\top X_s\right)^2$. This process is nonnegative and defines as supermartingale since
\begin{align*}
	\bE\left[ M^\lambda_{t+1}\middle|\cF_t\right] &= \bE\left[ \exp\left(\lambda^\top S_{t+1} -\frac{\sigma^2}{2}\lVert \lambda\rVert^2_{H^0_{t+1}(\theta^*)}\right) \middle|\cF_t\right]\\
	&=\bE\left[ \exp\left(\lambda^\top S_{t} -\frac{\sigma^2}{2}\lVert \lambda\rVert^2_{H^0_{t}(\theta^*)} + \partial\cL\left(Y_t, \langle\theta^*, X_t\rangle\right)\lambda^\top X_t - \frac{\sigma^2}{2}\partial^2\cL\left(Y_t, \langle\theta^*, X_t\rangle\right)\left(\lambda^\top X_t\right)^2 \right) \middle|\cF_t\right]\\
	&= \exp\left(\lambda^\top S_{t} -\frac{\sigma^2}{2}\lVert \lambda\rVert^2_{H^0_{t}(\theta^*)}\right) \bE\left[ \exp\left(\partial\cL\left(Y_t, \langle\theta^*, X_t\rangle\right)\lambda^\top X_t - \frac{\sigma^2}{2}\partial^2\cL\left(Y_t, \langle\theta^*, X_t\rangle\right)\left(\lambda^\top X_t\right)^2 \right) \middle|\cF_t\right]\\
	&\leq \exp\left(\lambda^\top S_{t} -\frac{\sigma^2}{2}\lVert \lambda\rVert^2_{H^0_{t}(\theta^*)}\right) \quad\text{(Lemma~\ref{lem:supermart})}\\
	&= M^{\lambda}_t\,.
\end{align*}

Now we construct a new supermartingale by mixing all the $M^\lambda$. More formally, let $\Lambda$ a $\R^d$-valued random variable independent of the rest and define $M_t=\bE\left[M^\Lambda_t \middle| \cF_{\infty}\right]$ where $\cF_{\infty} = \sigma\left(\bigcup\limits_{t\in\N}\cF_t\right)$. If $\Lambda$ has density $p$ with respect to the Lebesgue measure, this means that $M_t = \int_{\R^d} M^\lambda_t p(\lambda)d\lambda$. For the choice $\Lambda\sim \cN(0, \frac{1}{\beta\sigma^2}I_d)$ with $\beta>0$, we have, by completing the square in the exponential:
\begin{align*}
	M_t &= \frac{(\beta\sigma^2)^{d/2}}{(2\pi)^{d/2}}\int_{\R^d} \exp\left(-\lambda^\top S_t +\frac{\sigma^2}{2}\left(\lambda^\top\left(H^0_t(\theta^*) + \beta I_d\right)\lambda\right)\right)d\lambda\\
	&= \frac{(\beta\sigma^2)^{d/2}}{(2\pi)^{d/2}} \exp\left(\frac{\sigma^2}{2}\bar{\lambda}^\top H^\beta_t(\theta^*) \bar{\lambda}\right) \int_{\R^d} \exp\left(-\frac{\sigma^2}{2}\left(\lambda-\bar{\lambda}\right)^\top H^\beta_t(\theta^*) \left(\lambda-\bar{\lambda}\right)\right)d\lambda\\
	&= \left(\frac{\beta^d}{\det H_t^\beta(\theta^*)}\right)^{\frac{1}{2}} \exp\left(\frac{\sigma^2}{2}\bar{\lambda}^\top H^\beta_t(\theta^*) \bar{\lambda}\right)\,,
\end{align*}
where $\bar{\lambda} = \frac{1}{\sigma^2}H^\beta_t(\theta^*)^{-1}S_t$ and $H^\beta_t(\theta) = H^0_t(\theta) + \beta I_d$ is the regularized Hessian, which is positive definite and hence invertible. This expression further simplifies to $M_t=\left(\frac{\det \beta I_d}{\det H_t^\beta(\theta^*)}\right)^{\frac{1}{2}} \exp\left(\frac{1}{2\sigma^2}\lVert S_t \rVert^2_{H^\beta_t(\theta^*)^{-1}}\right)$.

From there, the argument is standard: $M^\lambda$ is a nonnegative supermartingale, and therefore the pointwise limit $M^\lambda_{\infty} = \lim_{t\rightarrow+\infty}M^\lambda_t$ exists almost surely (Doob's supermartingale convergence theorem, Ch. 11 in \citep{williams1991probability}). Therefore for any $\cF$-stopping time $\tau$, $M^\lambda_\tau$ is well-defined, and thus so is and $M_\tau $. By Fatou's lemma and Doob's stopping theorem, we have that $\bE[M_\tau]= \bE[\liminf_{t\rightarrow+\infty}\bE[M_{t\wedge\tau}\mid\cF_{\infty}]]\leq \liminf_{t\rightarrow+\infty}\bE[\bE[M_{t\wedge\tau}\mid\cF_{\infty}]]\leq 1$. Finally, the particular choice of $\tau=\inf\left\lbrace t\in\N,\ \lVert S_t \rVert^2_{H^\beta_t(\theta^*)^{-1}} \geq \sigma^2\left( 2\log\frac{1}{\delta} + \log \frac{\det \beta I_d}{\det H_t^\beta(\theta^*)}\right) \right\rbrace$ and a straightforward application of Markov's inequality reveals that
\[\bP\left(\tau < \infty\right) = \bP\left(\exists t\in\N,\ M_\tau \geq \frac{1}{\delta}\right)\leq \bE[M_\tau]\delta \leq \delta \,,\]
which is exactly the expected result.
$\hfill\blacksquare$
}

\section{ASSUMPTION~\ref{ass:loss_curvature} $\implies$ LEMMA~\ref{lem:transport}}\label{app:C}
	
First, we recall the two assumptions of interest.

\assumptionlosscurvature*

\lemmatransport*

Now simple calculations show that:

\begin{align*}\bar{H}^{\alpha}_t(\theta^*, \bar{\theta}_t) &= \sum_{s=1}^{t-1}\int_{0}^1 \partial^{2}\cL(Y_s, \langle u\theta^* + (1-u)\bar{\theta}_t, X_s\rangle)du X_s X_s^\top + \alpha I_d \\
&= \sum_{s=1}^{t-1}\int_{0}^1 \partial^{2}\cL(Y_s, \langle \theta^*, X_s\rangle)\frac{\partial^{2}\cL(Y_s, \langle u\theta^* + (1-u)\bar{\theta}_t, X_s\rangle)}{\partial^{2}\cL(Y_s, \langle \theta^*, X_s\rangle)}du X_s X_s^\top + \alpha I_d\\
&\succcurlyeq \frac{m}{M} \sum_{s=1}^{t-1} \partial^{2}\cL(Y_s, \langle \theta^*, X_s\rangle)X_s X_s^\top + \alpha I_d\\
&= \frac{1}{\kappa}\left(\sum_{s=1}^{t-1} \partial^{2}\cL(Y_s, \langle \theta^*, X_s\rangle)X_s X_s^\top + \kappa\alpha I_d\right)\\
&= \frac{1}{\kappa}H^{\kappa \alpha}_t(\theta^*)\,,
\end{align*}
which is the desired result if $\beta=\kappa\alpha$. The other inequality with $\bar{H}^{\beta}_t(\bar{\theta}_t)$ is derived similarly.

\section{PROOF OF THEOREM~\ref{thm:regret_ucb}}\label{app:proof_linucb}

In this section, we prove the main regret theorem for LinUCB with convex risk, which we restate below.

\theoremone*

\proof{We will prove the regret bound in two steps. First, we justify the choice of exploration sequence $(\gamma_t)_{t\in\N}$, which naturally derives from the optimistic principle and the analysis of local metrics. Then, we use a somewhat crude bound on the Hessian to simplify the analysis and reduce it to the so-called elliptic potential lemma.

Indeed, as established in Section~\ref{subsec:optimism_local_metrics}, the cumulative regret up to time $T$, denoted by $\cR_T$, is upper bounded with probability at least $1-\delta$ by $2\sum_{t=1}^T \gamma_t(X_t)$ provided that $\bP\left(\forall t\leq T,\ \Delta(X_t, \bar{\theta}_t) \leq \gamma_t(X_t) \right) \geq 1 - \delta$, where \[\Delta(X_t, \theta) = \lvert \langle \theta^*-\theta, X_t\rangle\rvert \leq \lVert \theta^* - \bar{\theta}_t\rVert_{\bar{H}^{\alpha}_t(\theta^*, \bar{\theta}_t)} \lVert X_t \rVert_{\bar{H}^{\alpha}_t(\theta^*, \bar{\theta}_t)^{-1}}\,.\]

\paragraph{Tuning of the Exploration Bonus Sequence}
The transportation of local metrics (Lemma~\ref{lem:transport}, implied by the curvature bound of Assumption~\ref{ass:loss_curvature}) reveals that 
\begin{align*}
\lVert \theta^* - \bar{\theta}_t\rVert_{\bar{H}^{\alpha}_t(\theta^*, \bar{\theta}_t)} &\leq \lVert F^{\alpha}_t(\theta^*) - F^{\alpha}_t(\wh{\theta}_t)\rVert_{\bar{H}^{\alpha}_t(\theta^*, \bar{\theta}_t)^{-1}} + \lVert  F^{\alpha}_t(\bar{\theta}_t) - F^{\alpha}_t(\wh{\theta})\rVert_{\bar{H}^{\alpha}_t(\theta^*, \bar{\theta}_t)^{-1}}\\
&\leq \sqrt{\kappa}\left(\lVert F^{\alpha}_t(\theta^*) - F^{\alpha}_t(\wh{\theta}_t)\rVert_{H^{\beta}_t(\theta^*)^{-1}} + \lVert  F^{\alpha}_t(\bar{\theta}_t) - F^{\alpha}_t(\wh{\theta})\rVert_{H^{\beta}_t(\bar{\theta}_t)^{-1}}\right)\,.
\end{align*}

Thanks to the supermartingale control of Lemma~\ref{lem:supermart}, we deduce from Corollary~\ref{corr:conf} that with probability at least $1-\delta$, the following inequalities hold for all $t\leq T$:
\begin{align*}
&\lVert F^{\alpha}_t(\theta^*) - F^{\alpha}_t(\wh{\theta}_t)\rVert_{H^{\beta}_t(\theta^*)^{-1}} \leq \sigma\sqrt{2\log\frac{1}{\delta} + \log\frac{\det H^\beta_t(\theta^*)}{\det \beta I_d}} + \alpha\lVert \theta^*\rVert_{H^\beta_t(\theta^*)^{-1}}\,,\\
&\lVert F^{\alpha}_t(\bar{\theta}_t) - F^{\alpha}_t(\wh{\theta}_t)\rVert_{H^{\beta}_t(\bar{\theta}_t)^{-1}} \leq \sigma\sqrt{2\log\frac{1}{\delta} + \log\frac{\det H^\beta_t(\bar{\theta}_t)}{\det \beta I_d}} + \alpha\lVert \bar{\theta}_t\rVert_{H^\beta_t(\bar{\theta}_t)^{-1}}\,.
\end{align*}
The prior bound on parameters (Assumption~\ref{ass:4}) yields $\lVert \theta\rVert_{H^{\beta}_t(\theta)^{-1}}\leq \frac{S}{\sqrt{\beta}}$ for $\theta\in\lbrace \theta^*, \bar{\theta}_t\rbrace$. Furthermore, the curvature bound (Assumption~\ref{ass:loss_curvature}) implies that $H^{\beta}_t(\theta) \preccurlyeq M V^{\beta/M}_t$, and therefore $\det H^{\beta}_t(\theta) \leq M^d \det V^{\beta/M}_t$ for $\theta\in\lbrace \theta^*, \bar{\theta}_t\rbrace$. Combining this together and substituting the expression of $\beta=\kappa \alpha$, where $\kappa=\frac{M}{m}$ is the conditioning of the convex loss $\cL$, we obtain:
\begin{align*}
\lVert \theta^* - \bar{\theta}_t\rVert_{\bar{H}^{\alpha}_t(\theta^*, \bar{\theta}_t)} &\leq 2\sqrt{\kappa}\left(\sigma\sqrt{2\log\frac{1}{\delta} + d\log\frac{m}{\alpha} + \log\det V^{\frac{\alpha}{m}}_t} + \sqrt{\frac{\alpha}{\kappa}}S\right)\,.
\end{align*}

By the same arguments, it holds that $\bar{H}^{\alpha}_t(\theta^*, \bar{\theta}_t)^{-1}\preccurlyeq \kappa H^{\kappa\alpha}_t(\bar{\theta}_t)^{-1}$ and therefore $\lVert X_t \rVert_{\bar{H}^{\alpha}_t(\theta^*, \bar{\theta}_t)^{-1}} \leq \sqrt{\kappa} \lVert X_t \rVert_{H^{\kappa\alpha}_t(\bar{\theta}_t)^{-1}}$. This shows that 
\[\gamma_t\colon x\in\cX_t \mapsto \underbrace{2\kappa\left(\sigma\sqrt{2\log\frac{1}{\delta} + d\log\frac{m}{\alpha} + \log\det V^{\frac{\alpha}{m}}_t} + \sqrt{\frac{\alpha}{\kappa}}S\right)}_{\eqqcolon\ c^\delta_t} \lVert x \rVert_{H^{\kappa\alpha}_t(\bar{\theta}_t)^{-1}}\]
is a valid choice of exploration sequence.

\paragraph{Bounding the Regret} Going back to the cumulative regret $\cR_T$, we notice that $(c^\delta_t)_{t=1,\dots, T}$ is a positive, nondecreasing sequence, therefore we have with probability at least $1-\delta$ that
\begin{align*}
	\cR_T \leq 2\sum_{t=1}^T \gamma_t(X_t) \leq 2 c^\delta_T \sum_{t=1}^T \lVert X_t \rVert_{H^{\kappa\alpha}_t(\bar{\theta}_t)^{-1}}\,.
\end{align*}
A priori, the direct analysis of the right-hand side is tedious due to the dependency on $\bar{\theta}_t$ in the local metric. However, we notice that the curvature bound (Assumption~\ref{ass:loss_curvature}) also implies the weaker control $H^{\kappa\alpha}_t(\bar{\theta}_t)^{-1} \preccurlyeq \frac{1}{m} (V^{\frac{\kappa \alpha}{m}}_t)^{-1}$, which translates to $\lVert X_t \rVert_{H^{\kappa\alpha}_t(\bar{\theta}_t)^{-1}} \leq \frac{1}{\sqrt{m}} \lVert X_t \rVert_{(V^{\frac{\kappa \alpha}{m}}_t)^{-1}}$. This bound is less informative as it looses the local information carried by $\bar{\theta}_t$, but still sufficient to obtain sublinear regret growth. We recall the following result, which is a direct consequence of the deterministic elliptic potential lemma (Lemma~11, \citet{abbasi_yadkori2011}) and the Cauchy-Schwarz inequality.

\begin{lemma}[Deterministic elliptic potential]\label{lemma:elliptical_potential_revisited}
Let $(x_t)_{t\in\N}$ denote an arbitrary sequence of vectors in $\cB^d_{\lVert \cdot \rVert}(0, L)$, $\epsilon > 0$ and $v_{t} = \sum_{s=1}^{t-1} x_s x_s^\top + \epsilon I_d \in \cS_d(\bR)$ for $t\in \N$. Then
\[\sum_{s=1}^t \lVert x_s \rVert_{v^{-1}_s} \leq \max(1, \frac{L}{\sqrt{\epsilon}})\sqrt{2td\log\left(\!1\!+\!\frac{t L ^2}{d\epsilon}\right)}\,.\]
\end{lemma}

Note that this result holds in our case (with $\epsilon=\frac{\kappa \alpha}{m}$) thanks to the prior bound on actions (Assumption~\ref{ass:5}).

\paragraph{Conclusion} With high probability, the regret of LinUCB with convex risk is bounded by
\begin{align*}
\cR_T \leq 2\sum_{t=1}^T \gamma_t(X_t) \leq 2c^\delta_T\max\left(\frac{1}{\sqrt{m}}, \frac{L}{\sqrt{\kappa \alpha}}\right) \sqrt{2Td\log\left(\!1\!+\!\frac{m ¸T L ^2}{d\kappa\alpha}\right)}\,.
\end{align*}
Going back to the expression of $c^\delta_T$, it follows from simple algebra (see e.g., \citet[proof of Lemma 19.4]{lattimore2020bandit}) that $\det V^\frac{\alpha}{m}_t \leq \left(\frac{\alpha}{m} + \frac{TL^2}{d}\right)^d$, and thus $c^\delta_T=\cO\left(\kappa\sigma \sqrt{d\log \frac{T L^2}{d}}\right)$ when $T\rightarrow +\infty$. A simpler asymptotic bound on the regret is therefore
\[\cR_T = \cO\left(\frac{\kappa\sigma d}{\sqrt{m}}\sqrt{T} \log \frac{T L^2}{d}\right)\,.\]
$\hfill\blacksquare$
}

%
%

\paragraph{Impact of using the local Hessian metric $H_t$ versus the global metric $V_t$}
	
To highlight the benefit of using local metrics, we detail here the regret bound obtained using the above proof with the natural global metric induced by $V^{\alpha/m}$ (independent of the local point $\theta$). Instantiating the positive definite matrix $P$ to $V^{\alpha/m}_t$ instead of $\bar{H}^{\alpha}(\theta^*, \bar{\theta}_t)$ in the bound on the prediction error of Section~\ref{subsec:optimism_local_metrics} yields
\begin{align*}
	\Delta(x, \bar{\theta}_t) &\leq \lVert \theta^* - \bar{\theta}_t\rVert_{V^{\alpha/m}_t} \lVert x\rVert_{(V^{\alpha/m}_t)^{-1}}\\
	&= \lVert F^{\alpha}_t(\theta^*) - F^{\alpha}_t(\bar{\theta}_t) \rVert_{\bar{H}^{\alpha}_t(\theta^*, \bar{\theta}_t)^{-1} V^{\alpha/m}_t \bar{H}^{\alpha}_t(\theta^*, \bar{\theta}_t)^{-1}} \lVert x\rVert_{(V^{\alpha/m}_t)^{-1}}\\
	&\leq \frac{1}{\sqrt{m}} \lVert F^{\alpha}_t(\theta^*) - F^{\alpha}_t(\bar{\theta}_t) \rVert_{\bar{H}^{\alpha}_t(\theta^*, \bar{\theta}_t)^{-1}} \lVert x\rVert_{(V^{\alpha/m}_t)^{-1}}\\
	&\leq 2\sqrt{\frac{\kappa}{m}} \left(\sigma\sqrt{2\log\frac{1}{\delta} + d\log\frac{m}{\alpha} + \log\det V^{\frac{\alpha}{m}}_t} + \sqrt{\frac{\alpha}{\kappa}}S\right) \lVert x\rVert_{(V^{\alpha/m}_t)^{-1}}\,.
\end{align*}	
Similarly to the above proof, this shows that
\[\gamma^{\text{global}}_t\colon x\in\cX_t \mapsto 2\sqrt{\frac{\kappa}{m}} \left(\sigma\sqrt{2\log\frac{1}{\delta} + d\log\frac{m}{\alpha} + \log\det V^{\frac{\alpha}{m}}_t} + \sqrt{\frac{\alpha}{\kappa}}S\right) \lVert x\rVert_{(V^{\alpha/m}_t)^{-1}}\]
is also a valid choice of exploration sequence. Finally, a straightforward application of Lemma~\ref{lemma:elliptical_potential_revisited} shows the regret of the corresponding LinUCB-CR strategy is upper bounded with probability at least $1 - \delta$ by
\[4\sqrt{\frac{\kappa}{m}} \left(\sigma\sqrt{2\log\frac{1}{\delta} + d\log\frac{m}{\alpha} + \log\det V^{\frac{\alpha}{m}}_t} + \sqrt{\frac{\alpha}{\kappa}}S\right) \sqrt{2Td \log\left(1+\frac{mTL^2}{d \alpha}\right)}\,.\]

Compared to the local analysis, this improves the scaling of the regret in $\kappa$ by a factor $\sqrt{\kappa}$. However, it forces the use of the global metric $\lVert \cdot \rVert_{(V^{\alpha/m}_t)^{-1}}$ instead of the local one $\lVert \cdot \rVert_{H^{\kappa\alpha}_t(\bar \theta_t)^{-1}}$, thus ignoring the precise shape of the loss function $\cL$. 

Looking at our proof, we see that $\kappa$ and $m$ fulfil two different roles. $\sqrt{\kappa}$ is the price to pay in order to transport local metrics $H^{\alpha}_t(\theta)$ between $\theta=\theta^*$ (true parameter) and $\theta=\bar{\theta}_t$ (estimate); it is paid once to bound the prediction error $\Delta(X_t, \theta)$ using the concentration bound of Proposition~\ref{prop:conf}, and it is also paid a second time if local metrics are used in the exploration bonus when moving from $\bar{H}_t(\theta^*, \bar{\theta}_t)^{-1}$ to $\bar{H}_t(\bar{\theta}_t)^{-1}$ (the former cannot be used directly in the algorithm as it depends on the a priori unknown paramter $\theta^*$). On the other hand, $m^{-1/2}$ is the price paid in both the local and global analyses to move from $H_t(\bar{\theta}_t)^{-1}$ to $V_t^{-1}$ in order to apply the elliptic potential lemma, which is in general incompatible with local metrics. A similar phenomenon is observed in the analysis of \citet{faury2020improved}: the regret of their algorithm Logistic-UCB-1 (global) scales as $\sqrt{\kappa}$ while that of Logistic-UCB-2 (local) scales as $\kappa$. In addition to local metrics, Logistic-UCB-2 also makes use of an intricate projection step that allows for a new elliptic potential lemma compatible with local metrics, thus removing the factor $m^{-1}$ (at least from the first order contribution to the regret in $T$). We conjecture that a similar analysis could be unlocked in the present risk-aware setting and leave it open for future investigation.

We reiterate that in the logistic setting, $\kappa$ is derived from self-concordance properties of the link function and is in particular independent of the curvature lower bound represented by $m$ (it is in fact equal to $1+2S$ where $S$ is an upper bound on the parameter space $\Theta$, as in Assumption~\ref{ass:4}). 
 By analogy with logistic bandits, we argue that the exact scaling in $\kappa$ is likely not too harmful for the practical performances of Algorithm~\ref{algo::UCB_convex_risk} and we therefore recommend the use of local metrics instead.
\section{PROOF OF THEOREM~\ref{thm:regret_ucb_bis}}\label{app:proof_ucb_bis}

In this section, we prove the regret bound of Theorem~\ref{thm:regret_ucb_bis} in the stochastic i.i.d. actions setting of Assumption~\ref{ass:6}. We first state the full regret bound with an explicit higher order term.

\begin{restatable}[Regret of LinUCB-CR with stochastic actions]{thm}{theoremonebis_full}\label{thm:regret_ucb_bis_full}
	Let $\delta\in(0, 1)$ and $t_0=\lceil \frac{8}{\rho_{\cX}^2}\log\frac{2}{\delta} - \frac{2\beta}{m\rho_{\cX}L^2}\rceil$. Under Assumptions~\ref{ass:loss_curvature}-\ref{ass:subgauss}-\ref{ass:4}-\ref{ass:5}-\ref{ass:6}, for $T\geq t_0$, with probability at least $1-2\delta$, the regret of Algorithm~\ref{algo::UCB_convex_risk} is bounded by
	\[\cR_T \leq 4 c^\delta_T \sqrt{\frac{2T}{m \rho_{\cX}}} \left(1 + \frac{C}{\sqrt{T}} \right)\,, \]
	where
	\[C=\frac{1}{L^2}\sqrt{\frac{\kappa \alpha - 4\frac{mL^2}{\rho_{\cX}} \log\frac{2}{\delta}}{2m\rho_{\cX}}} - \frac{1}{2L}\sqrt{t_0-1 + \frac{2(\kappa \alpha-\frac{4mL^2}{\rho_{\cX}}\log\frac{2}{\delta})}{m\rho_{\cX}L^2}} + \frac{1}{2}\max\left(1, L\sqrt{\frac{m}{\kappa \alpha}}\right)\sqrt{\rho_{\cX} d t_0 \log\left(1 + \frac{mL^2 t_0}{d\kappa \alpha}\right)}\,.\]
	
	In particular, we have $\cR_T = \cO\left(\kappa\sigma \sqrt{\frac{ dT}{m\rho_{\cX}}\log \frac{TL^2}{d}}\right)$.
\end{restatable}

The main difference with the proof of Theorem~\ref{thm:regret_ucb} is the use of an alternative stochastic elliptic potential lemma, mirroring the classical result of Lemma~\ref{lemma:elliptical_potential_revisited}, that exploits the lower bound on the covariance of actions (Assumption~\ref{ass:6}). This proof technique is adapted from \citet{kim2022double}, although we use a different, sharper concentration result (Proposition~\ref{prop:hilbert_mart_concentration} below).

\begin{restatable}[Stochastic elliptic potential lemma]{lem}{sto_ell_pot}\label{lem:sto_ell_pot}
	Let $\beta>0$, $\left(\theta_t\right)_{t\in\bN}$ a sequence of vectors in $\Theta\subseteq \bR^d$ and $\left(X_t\right)_{t\in\bN}$ a sequence of random variables in $\bR^d$. Recall that $H^{\beta}_t(\theta_t) = \sum\limits_{s=1}^{t-1}\partial^2\cL(Y_s, \langle X_s, \theta_t\rangle)X_s X_s^\top\!+\!\beta I_d$ and $V^{\beta}_t = \sum\limits_{s=1}^{t-1} X_s X_s^\top + \beta I_d$ for $t\in\bN$. Under Assumptions~\ref{ass:5} and \ref{ass:6}, let $\delta\in(0, 1)$ and $t_0=\lceil \frac{8}{\rho_{\cX}^2}\log\frac{2}{\delta} - \frac{2\beta}{m\rho_{\cX}L^2}\rceil$. For $T\geq t_0$, with probability at least $1-\delta$, it holds that
	\[\sum_{t=1}^T \lVert X_t\rVert_{H^{\beta}_t(\theta_t)^{-1}} \leq 2\sqrt{\frac{2T}{m\rho_{\cX}}}\left(1 + \frac{C}{\sqrt{T}}\right)\,,\]
	where 
	\[C=\frac{1}{L^2}\sqrt{\frac{\beta - 4\frac{mL^2}{\rho_{\cX}} \log\frac{2}{\delta}}{2m\rho_{\cX}}} - \frac{1}{2L}\sqrt{t_0-1 + \frac{2(\beta-\frac{4mL^2}{\rho_{\cX}}\log\frac{2}{\delta})}{m\rho_{\cX}L^2}} + \frac{\sqrt{\rho_{\cX}}}{2}\max\left(1, L\sqrt{\frac{m}{\beta}}\right)\sqrt{t_0 d \log\left(1 + \frac{mL^2 t_0}{d\beta}\right)}\,.\]
\end{restatable}

The intuition about this result is the following: if the matrix norms induced by $H^{\beta}_t(\theta_t)$ grow at least linearly in $t$, then the left-hand side should scale like $\sum_{t=1}^T \frac{1}{\sqrt{t}} = \cO(\sqrt{T})$, without the extra $\cO(\sqrt{\log T})$ factor present in Lemma~\ref{lemma:elliptical_potential_revisited}. The lower curvature bound of Assumption~\ref{ass:loss_curvature} shows that it is enough to look at the norms induced by $V^{\beta/m}_t$, at the cost of an extra $m^{-1/2}$ factor (in particular Lemma~\ref{lem:sto_ell_pot} holds for \textit{any} sequence $(\theta_t)_{t\in\bN}$, not just the sequence of estimators used in the bandit algorithms). Because of the stochastic sampling of actions (Assumption~\ref{ass:6}), it is likely that the sequence $(X_t)_{t\in\bN}$ spans all directions of $\bR^d$ quite fast; in other words, each new $X_t X_t^{\top}$ will contribute at least a fixed amount to the sum that defines $V^{\beta/m}_t$, leading to the linear growth of the induced norms.

We formalize this intuition in Lemma~\ref{lem:smallest_eig} below, which relies on the following concentration bound in Hilbert spaces.

\begin{restatable}[Time-uniform line crossing inequality for martingales with bounded increments in a Hilbert space]{prop}{hilbert_mart_concentration}\label{prop:hilbert_mart_concentration}
	Let $\left(\cH, \langle\cdot, \cdot\rangle\right)$ a Hilbert space and $\left(M_t\right)_{\in\bN}$ a $\cH$-valued martingale (with respect to a filtration $\left(\cF_t\right)_{t\in\bN}$) such that $M_0=0$. Assume that there exists a sequence of positive scalars $(c_t)_{t\in\bN}$ such that $\lVert M_{t+1} - M_{t}\rVert \leq c_t$ for all $t\in\bN$,
	where $\lVert \cdot \rVert$ denotes the norm induced by the scalar product. Then for any $\eta>0$ and $\delta\in(0, 1)$, it holds that 
	\[\bP\left( \exists t\in\bN,\ \lVert M_t \rVert \geq \frac{1}{2\eta}\log\frac{2}{\delta} + \eta \sum_{s=1}^t c_s^2 \right) \leq \delta\,.\]	 
\end{restatable}

Interestingly, this concentration bound does not depend on the dimension of $\cH$, and in particular remains valid even if the ambient space is infinite-dimensional. Moreover, this bound controls the probability that \textit{any} deviation occurs in the sequence $(\lVert M_t\rVert)_{t\in\bN}$, which is much stronger than controlling the deviation probability individually at each time $t\in\bN$. The proof relies on martingale arguments rather than a crude union bound over a finite set of individual deviation probabilities, which yields anytime ($t\in\bN$ rather than $t\leq T$ for some known horizon $T$) and typically tighter bounds.

\proof{
	This result is directly taken from \citet{howard2020time}. More precisely, \citet[Theorem~1]{howard2020time} shows a variety of equivalent time-uniform line crossing inequalities for martingales, and \citet[Corollary~10]{howard2020time} applies this generic result to concentration of norm-like operators in Banach spaces. In order to get the most convenient form for our problem, we derive Proposition~\ref{prop:hilbert_mart_concentration} from the generic theorem rather than the specific corollary.
	
The proofs of \citet[Theorem~3]{pinelis1992approach} and \citet[Theorem~3]{pinelis1994optimum} reveals that for any $\lambda\in\bR$, the exponential process $L_t=\cosh\left(\lambda \lVert M_t\rVert\right) \exp(-\frac{\lambda^2}{2}\sum\limits_{s=1}^t c_s)$ is a nonnegative $\cF$-supermartingale. Therefore, \citet[Theorem~1,~(a)]{howard2020time} shows that for any $a,b>0$,
\[\bP\left( \exists t\in\bN,\ S_t \geq a + bV_t \right) \leq 2e^{-aD(b)}\,,\]
where $S_t=\lVert M_t\rVert$, $V_t=\sum\limits_{s=1}^t c_s^2$ and $D(b)=2b$. Equating the right-hand side to $\delta$ and letting $b=\eta$ concludes the proof, with $a=\frac{1}{2\eta}\log\frac{2}{\delta}$.
$\hfill\blacksquare$
}

In the next lemma, we show that the smallest eigenvalue of $H^{\beta}_t(\theta_t)$, which provides a lower bound to the corresponding induced norm, does indeed grow linearly with $t$ on an event of high probability. For a given symmetric matrix $A\in\cS_d(R)$, we denote by $\lambda_{\min}(A)$ its smallest eigenvalue. 

\begin{restatable}[Smallest eigenvalue of $H^{\beta}_t(\theta_t)$ grows linearly with $t$ with high probability]{lem}{lem_smallest_eig}\label{lem:smallest_eig}
	Under Assumptions~\ref{ass:loss_curvature}-\ref{ass:5}-\ref{ass:6}, it holds that
	\[\bP\left(\exists t\in\bN,\ \lambda_{\min}\left(H^\beta_{t+1}(\theta_{t+1})\right) \leq \beta - \frac{4mL^2}{\rho_{\cX}}\log\frac{2}{\delta} + \frac{m\rho_{\cX}L^2}{2} t\right) \leq \delta\,.\]
\end{restatable}

\proof{
First notice that Assumption~\ref{ass:loss_curvature} implies
\begin{align*}
	\lambda_{\min}\left(H^{\beta}_{t+1}(\theta_{t+1})\right) &\geq m \lambda_{\min}\left(V^0_{t+1}\right) + \beta\,.
\end{align*}
The idea is to relate $\lambda_{\min}\left(V^0_{t+1}\right)$ to the norm of some martingale in order to apply Proposition~\ref{prop:hilbert_mart_concentration}. In the stochastic actions setting (Assumption~\ref{ass:6}), a natural martingale is defined the following sum of random matrices:
\[M_t = \sum_{s=1}^t X_s X_s^\top - \bE\left[X_s X_s^\top \lvert \cF_{s-1} \right] = V^0_{t+1} - \bar{V}^0_{t+1}\,,\]
where we defined $\bar{V}^0_{t+1} = \sum\limits_{s=1}^t \bE\left[ X_{s}X_{s}^\top \lvert \cF_{s-1}\right]$. We recall that a consequence of Weyl's inequality on eigenvalues is that for any $A, B\in\cS_d(\bR)$, the following inequality holds:
\[\lambda_{\min}(A) + \lambda_{\min}(B) \leq \lambda_{\min}(A+B)\,.\]
Applying this to $A=M_t$ and $B=\bar{V}^0_{t+1}$ yields $\lambda_{\min}\left(M_t\right) + \lambda_{\min}\left(\bar{V}^0_{t+1}\right) \leq \lambda_{\min}\left(V^0_{t+1}\right)$. Now, notice that $\lambda_{\min}(A)=-\lambda_{\max}(A)\geq -\lVert A\rVert$ where $\lVert \cdot\rVert$ is the matrix norm induced by the scalar product $\langle A, B\rangle=\text{Tr}\left(A^\top B\right)$ (also known as the Frobenius norm). Moreover, the conditional covariance lower bound of Assumption~\ref{ass:6} and another application of Weyl's inequality imply that \[\lambda_{\min}\left(\bar{V}^0_{t+1}\right)\geq \sum_{s=1}^t \lambda_{\min}\left(\bE\left[ X_s X^\top_s \lvert \cF_{s-1}\right]\right) \geq \rho_{\cX}L^2 t\,.\] 

Combining these together, we obtain that for arbitrary $a\in\bR$ and $b>0$, the following inequality holds:
\begin{align}
\bP\left(\exists t\in\bN,\ \lambda_{\min}\left( H^\beta_{t+1}(\theta_{t+1})\right) \leq a + bt \right)\leq \bP\left(\exists t\in\bN,\ \lVert M_t \rVert  \geq \frac{\beta - a}{m} + \left(\rho_{\cX}L^2 - \frac{b}{m}\right)t \right)\,.
\end{align}
Notice that $\lVert M_{t+1} - M_{t}\rVert = \lVert X_t X_t^\top - \bE\left[X_t X_t^\top \lvert \cF_{t-1}\right]\rVert\leq 2 L^2$ (Assumption~\ref{ass:5}). Now if we choose $b=\frac{1}{2}m\rho_{\cX}L^2$ and $a=\beta - \frac{4mL^2}{\rho_{\cX}}\log\frac{2}{\delta}$, the bound from Proposition~\ref{prop:hilbert_mart_concentration} holds with $\eta=\frac{\rho_{\cX}}{8L^2}$ and $\delta\in(0, 1)$ and $c_t = 2L^2$, thus proving the result.

$\hfill\blacksquare$
}

We are now ready to prove the stochastic elliptic potential lemma.

\proof{{\bf of Lemma~\ref{lem:sto_ell_pot}} We fix $t_0\in\bN$ arbitrarily for now and let $T\geq t_0$. We start by splitting the sum in two and by applying the deterministic elliptic potential lemma (Lemma~\ref{lemma:elliptical_potential_revisited}) up to time $t_0$:
\begin{align*}
	\sum_{t=1}^T \lVert X_t \rVert_{H^{\beta}_t(\theta_t)^{-1}} &= \sum_{t=1}^{t_0} \lVert X_t \rVert_{H^{\beta}_t(\theta_t)^{-1}} + \sum_{t=t_0 + 1}^{T} \lVert X_t \rVert_{H^{\beta}_t(\theta_t)^{-1}}\\
	&\leq \frac{1}{\sqrt{m}}\sum_{t=1}^{t_0} \lVert X_t \rVert_{\left(V^{\beta/m}_t\right)^{-1}} + \sum_{t=t_0 + 1}^{T} \lVert X_t \rVert_{H^{\beta}_t(\theta_t)^{-1}} \text{\quad (Assumption~\ref{ass:loss_curvature})}\\
	&\leq \left(\frac{1}{\sqrt{m}}, \frac{L}{\sqrt{\beta}}\right)\sqrt{2t_0 d \log\left( 1 + \frac{mL^2 t_0}{d\beta}\right)} + \sum_{t=t_0}^{T-1} \lVert X_{t+1} \rVert_{H^{\beta}_{t+1}(\theta_{t+1})^{-1}} \text{\quad (Assumption~\ref{ass:5})}\,.
\end{align*}

Now let $\cE^{\delta}_t = \left\lbrace \forall t'\geq t,\ \lambda_{\min}\left(H^\beta_{t+1}(\theta_{t+1})\right) > \beta - \frac{4mL^2}{\rho_{\cX}}\log\frac{2}{\delta} + \frac{m\rho_{\cX}L^2}{2} t \right\rbrace$. It is clear that $\cE^{\delta}_{t_0}\subseteq \cE^{\delta}_{0}$, and thus by Lemma~\ref{lem:smallest_eig}, $\bP\left(\cE^{\delta}_{t_0}\right)\geq 1 - \delta$. The choice $t_0=\lceil \frac{8}{\rho_{\cX}^2}\log\frac{2}{\delta} - \frac{2\beta}{m\rho_{\cX}L^2}\rceil$ implies that the right-hand side in the definition of $\cE_{t_0}$ is positive. On this event, we bound the second sum as follows:
\begin{align*}
	\sum_{t=t_0}^{T-1} \lVert X_{t+1} \rVert_{H^{\beta}_{t+1}(\theta_{t+1})^{-1}} &\leq 
	L \sum_{t=t_0}^{T-1} \frac{1}{\sqrt{a + b t}}\\
	&\leq \frac{2L}{b}\left(\sqrt{a + b(T-1)} - \sqrt{a + b(t_0-1)}\right)\\
	&\leq \frac{2L}{b}\left(\sqrt{bT} + \sqrt{a} - \sqrt{a + b(t_0-1)}\right)\,,
\end{align*}
where we use the shorthand $a=\beta - \frac{4mL^2}{\rho_{\cX}}\log\frac{2}{\delta}$ and $b=\frac{1}{2}m\rho_{\cX}L^2$ (the penultimate line comes from sum-integral comparison while the last one follows from the inequality $\sqrt{x+y}\leq \sqrt{x} + \sqrt{y}$). After collecting the dominating term in $\sqrt{T}$, the two sums give the following upper bound:
\begin{align*}
\sum_{t=1}^{T} \lVert X_{t} \rVert_{H^{\beta}_{t}(\theta_t)^{-1}} &\leq 2L\sqrt{\frac{T}{b}}\left(1 + \frac{C}{\sqrt{T}} \right)
\end{align*}
with
\[C = \sqrt{\frac{a}{b}} - \sqrt{\frac{a+b (t_0-1)}{b}} + \frac{\sqrt{b}}{2L}\left(\frac{1}{\sqrt{m}}, \frac{L}{\sqrt{\beta}}\right)\sqrt{2t_0 d \log\left( 1 + \frac{mL^2 t_0}{d\beta}\right)}\,.\]
Substituting $a$ and $b$ with their expressions yields the result.

$\hfill\blacksquare$
}

We finally prove the regret bound in the stochastic i.i.d. actions setting.

\proof{{\bf of Theorem~\ref{thm:regret_ucb_bis_full}}
We follow the exact same steps as with Theorem~\ref{thm:regret_ucb} in order to bound the regret by
\[\cR_T \leq 2c^{\delta}_T \sum_{t=1}^T \lVert X_t \rVert_{H^{\kappa\alpha}_t(\bar{\theta}_t)}^{-1}\,, \]
with probability at least $1-\delta$. The sum on the left-hand side is controlled by Lemma~\ref{lem:sto_ell_pot} also with probability at least $1-\delta$. A simple union argument over both events concludes the proof, resulting in a regret upper bound with probability at least $1-2\delta$.

$\hfill\blacksquare$
}

We conclude this section with two remarks.

\begin{remark}[Dependency of $\rho_{\cX}$ on $d$]
	We recall that in the stochastic actions setting (Assumption~\ref{ass:6}), $\rho_{\cX}$ is a lower bound on the conditional covariance of actions, which can be equivalently formulated as $\rho_{\cX}L^2 \leq \lambda_{\min}\left(\bE\left[X_{t} X_{t}^\top \lvert \cF_{t-1}\right]\right)$ for all $t\in\bN$. Following the argument of \citet{kim2021doubly} in the case of unconditional covariance control, we obtain the following bound on $\rho_{\cX}$:
	\begin{align*}
		d\rho_{\cX}L^2 \leq d\lambda_{\min}\left(\bE\left[X_{t} X_{t}^\top \lvert \cF_{t-1}\right]\right) \leq \sum_{\lambda\in\cS_{t}} \lambda = {\normalfont\text{Tr}}\left(\bE\left[X_{t} X_{t}^\top \lvert \cF_{t-1}\right]\right) = \bE\left[{\normalfont\text{Tr}} \left(X_{t} X_{t}^\top\right) \lvert \cF_{t-1}\right] \leq L^2\,,
	\end{align*}
	where $\cS_{t}$ denotes the spectrum of the symmetric matrix $\bE\left[X_{t} X_{t}^\top \lvert \cF_{t-1}\right]$ and $\normalfont\text{Tr}$ the trace operator. Therefore $\rho_{\cX}\leq d^{-1}$. Moreover, \citet{bastani2021mostly, kim2021doubly, kim2022double} identified two families of examples where $\rho_{\cX}=\cO(d^{-1})$ in the unconditional case:
	\begin{itemize}
		\item If the distribution of $X\in\cX_t$ (marginal distribution of a each action) admits a density $p$ with respect to the Lebesgue measure supported in $\cB^d_{\lVert \cdot \rVert_2}(0, L)$ and such that $p(x)\geq p_{\min}>0$ for all $x\in\cB^d_{\lVert \cdot \rVert_2}(0, L)$, then $\rho_{\cX}=\frac{p_{\min}}{(d+2)} {\normalfont\text{vol}}\left(\cB^d_{\lVert \cdot \rVert_2}(0, 1)\right)$ is a suitable choice \citep[Lemma~C.1]{kim2022double}. In general, the volume of the Euclidean unit ball in $\bR^d$ is ${\normalfont\text{vol}}\left(\cB^d_{\lVert \cdot \rVert_2}(0, 1)\right) = \frac{\pi^{d/2}}{\Gamma(\frac{d}{2}+1)}\sim\frac{1}{\sqrt{d\pi}}\left(\frac{2\pi e}{d}\right)^\frac{d}{2}$, which goes to $0$ when $d\rightarrow +\infty$. In certain cases though, such as the uniform and truncated Gaussian distributions, $p_{\min}$ is proportional to ${\normalfont\text{vol}}\left(\cB^d_{\lVert \cdot \rVert_2}(0, 1)\right)$, thus leading to $\rho_{\cX}=\cO(d^{-1})$.
		\item If the covariance matrix $\bE\left[X X^\top\right]$ exhibits a certain structure, for instance  AR(1), tridiagonal or block diagonal, then $\rho_{\cX}=\cO(d^{-1})$, regardless of the marginal distributions.
	\end{itemize}
\end{remark}

\begin{remark}[Previous results about the growth of the smallest eigenvalue]
	\citet{kim2021doubly, kim2022double} prove similar results on the linear growth of the smallest eigenvalues of a different sequence of Hessian matrices. More precisely, they consider a fixed number $K$ of arms, i.e., action sets of the form $\cX_t=\left\lbrace X_{k, t}, k\in[K]\right\rbrace$ and Hessian matrices constructed from \textit{all} actions $V^{[K]}_{t+1} =\sum_{k\in[K]} \sum_{s=1}^t X_{k, s}X_{k, s}^{\top}$, instead of using only the actions played at previous time steps. This is made possible in their analyses by resorting to a doubly robust imputation of unobserved rewards associated to unplayed actions, which is significantly different from our approach. One theoretical benefit of their method is that the sequence $(V^{[K]}_{t+1})_{t\in\bN}$ can be more easily transformed into a $\cF$-martingale using only the unconditional lower bound on the covariance, as opposed to the conditional one of Assumption~\ref{ass:6}.
	
	Of note, \citet{li2017provably} also questions the feasibility of a linear lower bound on the smallest eigenvalue of $V^0_{t+1}$ but concludes that it requires more stringent assumption as \citet[Example~1]{lai1982least} seemingly provides a counterexample of sublinear growth in the context of a regression problem. However, this counterexample studies autoregressive actions instead of i.i.d. action sets, which leads to $\bE\left[X_t X_t^\top \lvert \cF_{t-1} \right] \rightarrow 0$ when $t\rightarrow +\infty$. Therefore, this is different from what we consider in Assumption~\ref{ass:6} and does not invalidate our analysis.
\end{remark}

%
\section{PROOFS OF OGD CONCENTRATION}\label{appendix:proof_sgd}

We reformulate below Proposition~\ref{prop:ogd} in full details and provide a proof of the OGD regret.

\begin{restatable}[OGD Regret, Sub-Gaussian Gradients]{prop}{propogd_full}\label{prop:ogd_full}
	Let $\cC$ a convex subset of $\R^d$ and $\Pi$ the projection operator onto $\cC$. For $j=1, \dots, N$, let $\ell_j\colon \cC \longrightarrow \R_+$ a twice differentiable convex function and $a, A >0$ such that $a I_d \preccurlyeq \nabla^2 \ell_j(z) \preccurlyeq A I_d$ for all $z\in\cC$. Define the OGD update at step $j$ by $z_j=\Pi(z_{j-1} - \epsilon_{j-1}\nabla \ell_j(z_{j-1}))$ and $\bar{z}_n=\arg\min_{z\in\cC} \sum_{j=1}^n \ell_j(z)$. Assume that there exists $z^*\in\cC$ such that $\nabla \ell_j(z^*) = g_j + \frac{\alpha}{n} z^*$ with $\alpha\geq 0$ and $g$ a centered, $\R^d$-valued $\sigma$-sub-Gaussian process, and also that $\cC$ is bounded, i.e $\text{diam}(\cC) = \sup_{z,z'\in\cC} \lVert z-z'\rVert<\infty$.
	Then with probability at least $1\!-\!\delta$, the OGD regret with step size $\epsilon_j\!=\!\frac{3}{aj}$ is bounded for all $n\leq N$ by
	\[\sum_{j=1}^n \ell_j(z_j) - \ell_j(\bar{z}_n) \leq \frac{9}{2a}\left(2d\sigma^2\log \frac{2dN}{\delta} + A^2 \text{diam}(\cC)^2 + \frac{\alpha^2}{n^2}\lVert z^*\rVert^2\right) \left(1 + \log n\right)\,.\]
	This can be written more concisely as $\sum_{j=1}^N \ell_j(z_j) - \ell_j(\bar{z}_N) = \cO(\frac{d\sigma^2}{a}\log^2 N)$ when $N\rightarrow +\infty$. In addition, if $g$ is uniformly bounded by a constant $G>0$, the regret with step size $\epsilon_s\!=\!\frac{1}{aj}$ can be reduced to the almost sure bound:
	\[ \sum_{j=1}^n \ell_j(z_j) - \ell_j(\bar{z}_n) \leq \frac{G^2}{2a}\left( 1+ \log n\right)\,.\]
\end{restatable}

\proof{
Let $j\leq n$. The uniform lower bound on the Hessian of $\ell_j$ makes it $a$-strongly convex, which implies
\[ \ell_j(z_j) - \ell_j(\bar{z}) \leq \langle\nabla \ell_j(z_j), z_j - \bar{z}_n \rangle - \frac{a}{2}\lVert \bar{z} - z_j\rVert^2. \]

By definition of the OGD scheme, the following holds:
\begin{align*} 
    \lVert z_{j+1} - \bar{z}_n \rVert^2 &= \lVert \Pi\left(z_j - \epsilon_j \nabla \ell_j(z_j)\right) - \bar{z}_n \rVert^2\\
    &\leq \lVert z_j - \epsilon_j \nabla \ell_j(z_j) - \bar{z}_n \rVert^2 \quad\text{(projection onto a convex set)}\\
    &\leq \lVert z_j - \bar{z}_n\rVert^2 + \epsilon_j^2 \lVert \nabla \ell_j(z_j) \rVert^2 - 2\epsilon_j \langle\nabla \ell_j(z_j), z_j - \bar{z}_n\rangle\,,\\
\end{align*}

from which we deduce

\[ \langle \nabla \ell_j(z_j), z_j - \bar{z}_n\rangle \leq \frac{\lVert z_j - \bar{z}_n \rVert^2 - \lVert z_{j+1} - \bar{z}_n \rVert^2}{2\epsilon_j} + \frac{\epsilon_j}{2}\lVert\nabla \ell_j(z_j) \rVert^2\,.\]

\paragraph{Bounded Gradients}\ \\
This case is covered by Theorem 3.3 in \citet{hazan2019introduction}. We reproduce the proof here for reference and as a first step toward the more general setting of sub-Gaussian gradients.

Let $G>0$ be such that $\lVert \nabla \ell_j(z_j)\rVert \leq G$ for all $j=1, \dots, n$. This allows to upper bound the above equation, leading to

\[\langle \nabla \ell_j(z_j), z_j - \bar{z}_n\rangle \leq \frac{\lVert z_j - \bar{z}_n \rVert^2 - \lVert z_{j+1} - \bar{z}_n \rVert^2}{2\epsilon_j} + \frac{\epsilon_j}{2}G^2.\]

The online regret of OGD is therefore

\[\sum_{j=1}^n \ell_j(z_j) - \ell_j(\bar{z}_n) \leq \frac{1}{2}\sum_{j=1}^n \frac{\lVert z_j - \bar{z}_n \rVert^2 - \lVert z_{j+1} - \bar{z}_n \rVert^2}{\epsilon_j} - a \lVert z_j - \bar{z}_n \rVert^2 + \frac{G^2}{2} \sum_{j=1}^n \epsilon_j. \]

The first sum can be rewritten after a simple index shift and the convention $1/\epsilon_0\coloneqq 0$:
\begin{align*}
\frac{1}{2}\sum_{j=1}^n \frac{\lVert z_j - \bar{z}_n \rVert^2 - \lVert z_{j+1} - \bar{z}_n \rVert^2}{\epsilon_j} - a \lVert z_j - \bar{z}_n \rVert^2 &= \frac{1}{2}\sum_{j=1}^n \lVert z_j - \bar{z}_n \rVert^2\left(\frac{1}{\epsilon_j} - \frac{1}{\epsilon_{j-1}} - a \right) - \frac{1}{\epsilon_n} \lVert z_{n+1} - \bar{z}_n \rVert^2\\
&\leq \frac{1}{2}\sum_{j=1}^n \lVert z_j - \bar{z}_n \rVert^2\left(\frac{1}{\epsilon_j} - \frac{1}{\epsilon_{j-1}} - a \right)\\
&= 0\\
\end{align*}
for the choice $\epsilon_j = \frac{1}{aj}$. Consequently, the online regret can be simplified as

\begin{align*}
    \sum_{j=1}^n \ell_j(z_j) - \ell_j(\bar{z}_n) &\leq \frac{G^2}{2} \sum_{j=1}^n \epsilon_j\\
    &=\frac{G^2}{2a} \sum_{j=1}^n \frac{1}{j}\\
    &\leq \frac{G^2}{2a}\left(1+\log n\right)\,.\\
\end{align*}

\paragraph{Sub-Gaussian Gradients}\ \\

We do not assume here that $\nabla \ell_j(z_j)$ is uniformly bounded, but instead rely on the weaker assumption that $\nabla \ell_j(z^*)$ is sub-Gaussian. The strategy is to control the variation between $\nabla \ell_j(z_j)$ and $\nabla \ell_j(z^*)$ on the one hand, and bound in high probability $\nabla \ell_j(z^*)$ on the other hand.

Notice that $\nabla \ell_j(z_j) = g_j + \frac{\alpha}{n} z^* + \nabla \ell_j(z_j) - \nabla \ell_j(z^*)$ and that there exists $\bar{z}_n\in [z_j, z^*]\subset \cC$ such that $\nabla \ell_j(z_j) - \nabla \ell_j(z^*) = \nabla^2 \ell_j(\bar{z}_n) \left(z_j - z^*\right)$ thanks to the mean value theorem and the convexity of $\cC$. This yields
\begin{align*}
    \lVert \nabla \ell_j(\phi_j)\rVert^2 &\leq 3\lVert g_j \rVert^2 + \frac{3\alpha^2}{n^2} \lVert z^* \rVert^2 + 3 \lVert \nabla \ell_j(z_j) - \nabla \ell_j(z^*)\rVert^2\\
    &\leq 3\lVert g_j \rVert^2 + \frac{3\alpha^2}{n^2} \lVert z^* \rVert^2 + 3 A^2 \lVert z_j - z^*\rVert^2\,,\\
\end{align*}
since $\nabla \ell_j$ is $A$-Lipschitz. Combining this with the above yields
\begin{align*}
    \langle \nabla \ell_j(z_j), z_j - \bar{z}_n\rangle &\leq \frac{3}{2}\frac{\lVert z_j - \bar{z}_n \rVert^2 - \lVert z_{j+1} - \bar{z}_n \rVert^2}{\epsilon_j} + \frac{3}{2}\epsilon_j \lVert g_j\rVert^2 + \frac{3}{2}\epsilon_j \frac{\alpha^2}{n^2} \lVert z^* \rVert^2 + \frac{3}{2}\epsilon_j A^2 \lVert z_j - z^* \rVert^2\\
    &\leq \frac{3}{2}\frac{\lVert z_j - \bar{z}_n \rVert^2 - \lVert z_{j+1} - \bar{z}_n \rVert^2}{\epsilon_j} + \frac{3}{2}\epsilon_j \left(\lVert g_j \rVert^2 + \frac{\alpha^2}{n^2} \lVert z^*\rVert^2 \right) + \frac{3}{2}\epsilon_j A^2 \text{diam}(\cC)^2 \,.\\
\end{align*}

The online regret of OGD is therefore

\begin{align*}\sum_{j=1}^n \ell_j(z_j) - \ell_j(\bar{z}_n) \leq &\frac{3}{2}\sum_{j=1}^n \frac{\lVert z_j - \bar{z}_n \rVert^2 - \lVert z_{j+1} - \bar{z}_n \rVert^2}{\epsilon_j} - \frac{a}{3} \lVert z_j - \bar{z}_n \rVert^2 \\
&+ \frac{3}{2} A^2 \text{diam}(\cC)^2 \sum_{j=1}^n\epsilon_j + \frac{3}{2}\sum_{j=1}^n \epsilon_j \left(\lVert g_j \rVert^2 + \frac{\alpha^2}{n^2} \lVert z^*\rVert^2 \right)\,.
\end{align*}

As in the bounded case, the choice $\epsilon_j = \frac{3}{aj}$ makes the first sum vanish. Moreover, a simple union argument over the Chernoff bound for the $\sigma$-sub-Gaussian random variables $(g_j)_{j=1, \dots, n}$ reveals that

\[ \cE_n = \left\lbrace \forall j=1, \dots, n,\ \lVert g_j \rVert \leq \sigma\sqrt{2d\log \frac{2dn}{\delta}} \right\rbrace \]
holds with probability at least $1 -\delta$ for $\delta\in(0, 1)$. Therefore, the following holds with probability at least $1-\delta$:

\begin{align*}
    \sum_{j=1}^n \epsilon_j \lVert g_j \rVert^2 \leq \sum_{j=1}^n \epsilon_j \lVert g_j \rVert^2 \ind_{\cE_n} \leq 2d\sigma^2\log \frac{2dn}{\delta}\sum_{j=1}^n \epsilon_j \,.
\end{align*}

Therefore, with probability at least $1-\delta$, we obtain the following online regret:
\begin{align*}
    \sum_{j=1}^n \ell_j(z_s) - \ell_j(\bar{z}_n) &\leq\frac{3}{2}\left(2d\sigma^2\log \frac{2dn}{\delta} + A^2 \text{diam}(\cC)^2 + \frac{\alpha^2}{n^2}\lVert z^*\rVert^2\right) \sum_{j=1}^n \epsilon_j\\
     &= \frac{9}{2a}\left(2d\sigma^2\log \frac{2dn}{\delta} + A^2 \text{diam}(\cC)^2 + \frac{\alpha^2}{n^2}\lVert z^*\rVert^2 \right) \sum_{j=1}^n \frac{1}{j}\\
     &\leq \frac{9}{2a}\left(2d\sigma^2\log \frac{2dn}{\delta} + A^2 \text{diam}(\cC)^2 + \frac{\alpha^2}{n^2}\lVert z^*\rVert^2\right) \left(1 + \log n\right)\,.
\end{align*}
$\hfill\blacksquare$
}

\section{PROOFS OF THEOREM~\ref{thm:regret_ucb-ogd}}\label{appendix:proof_ucb-ogd}

In this section, we adapt the regret analysis of LinUCB to the LinUCB-OGD variant that relies on online gradient approximation of the empirical risk minimizer.

\theoremtwo*

\proof{
Let $\ell_j(\theta) = \sum\limits_{k=1}^{h} \cL(Y_{(j-1)h+k}, \langle \theta, X_{(j-1)h+k}\rangle)+\frac{\alpha}{2N}\lVert \theta \rVert^2_2$, where $N=\lceil\frac{T-1}{h}\rceil$ denotes the total number of episodes of length $h$. For simplicity, we assume that $\bar{\theta}_t = \wh{\theta}_t$ for all $t\leq T$, i.e., the empirical risk minimizer is always in the stable set of the projection operator $\Pi$. We recall that $\sum_{j=1}^{n} \nabla \ell_j(\bar{\theta}_t) = 0$ for $n=\frac{t-1}{h}$ (i.e., after episode $n$, when $\wh{\theta}^{\text{OGD}}_n$ is updated). In the general case, replacing $\wh{\theta}_t$ by $\bar{\theta}_t$ induces an extra correction factor in the inequalities below which is at most polylogarithmic in $T$ (a consequence of Corollary~\ref{corr:conf}), and hence does not change the conclusion. Again, we point out that, similarly to the generalized linear bandit setting \citep{filippi2010parametric, faury2020improved}, $\wh{\theta}_t$ is often in the stable set of $\Pi$ in practice.

We use the notations of Proposion~\ref{prop:ogd} and define: 
\begin{align*}
	&z_j = \wh{\theta}^{\text{OGD}}_j\,,\\
	&\bar{z}_n = \bar{\theta}_{t}\,,\\
	&z^* = \theta^*\,.
\end{align*}

We also denote by $\tilde{z}_n = \bar{\theta}^{\text{OGD}}_n = \frac{1}{n}\sum_{j=1}^n z_j$ the average of the past $n$ OGD updates.

\paragraph{Bound on $\lVert \tilde{z}_{n} - \bar{z}_{n}\rVert_2$} 

Without loss of generality, we assume here that $\partial\cL^*$ is a $\sqrt{m}\sigma$ sub-Gaussian process (this follows in variety of settings from the discussion of Assumption~\ref{ass:subgauss} and Lemma~\ref{lem:supermart}; the conclusions are essentially unchanged when assuming only Assumption~\ref{ass:subgauss}, at the cost of slightly heavier notations).

We first note that $\nabla \ell_j(\theta^*) = g_j(\theta^*) + \frac{\alpha}{N}\theta^*$, where $g_j(\theta^*) = \sum_{k=1}^h \partial\cL^*_{(j-1)h+k}$ and $j\in[N]$, is $\sqrt{hm}\sigma$-sub-Gaussian (sum of $h$ random variables, each of them being drawn from a $\sqrt{m}\sigma$-sub-Gaussian distribution). Setting the episode length to $h=\lceil\frac{2\epsilon_h}{\rho_{\cX}L^2} + \frac{8}{\rho_{\cX}^2}\log\frac{2}{\delta}\rceil$ makes the one-step losses $\ell_j$ $m\epsilon_h$-strongly convex with high probability. Indeed, let us define for $h'\in\bN$ the function $f(h')=\frac{\rho_{\cX} L^2}{2}h' - \frac{4L^2}{\rho_{\cX}}\log\frac{2}{\delta}$ and the event $\cE_{j, h'} = \left\lbrace \lambda_{\min}\left(\sum\limits_{k=1}^h X_{(j-1)h'+k}X^\top_{(j-1)h'+k} \right) > f(h') \right\rbrace$. First, notice that $\cE_{j, h} \supseteq \bigcap\limits_{h'\in\bN} \cE_{j, h'}$ and that $\sum\limits_{k=1}^h X_{(j-1)h'+k}X^\top_{(j-1)h'+k}$ has the same distribution as $V^0_{h'+1}$ by Assumption~\ref{ass:6}. We deduce from Lemma~\ref{lem:smallest_eig} applied to $V^0_{h+1}$ (that is with $\beta=0$ and $m=1$), that
\begin{align*}
	\bP\left(\cE_{j, h}\right) \geq \bP(\bigcap\limits_{h'\in\bN} \cE_{j, h'})
	&= \bP\left(\forall h'\in\bN,\ \lambda_{\min}\left(V^0_{h'+1}\right) > - \frac{4L^2}{\rho_{\cX}}\log\frac{2}{\delta} + \frac{\rho_{\cX}L^2}{2} h'\right) \geq 1 - \delta\,.
\end{align*}
In particular for the value of $h$ defined above, we have $\cE_{j, h}\subseteq \left\lbrace \lambda_{\min}\left(\sum\limits_{k=1}^h X_{(j-1)h'+k}X^\top_{(j-1)h'+k}\right) \geq \epsilon_h \right\rbrace$, which gives the $m\epsilon_h$-strong convexity of $\ell_j$ by the usual minoration $\partial^2\cL \geq m$ (Assumption~\ref{ass:loss_curvature}). In the rest of this proof, we assume to be on the event $\bigcap\limits_{j\in[N]} \cE_{j,h}$, the probability of which is at least $1-N\delta$ by a simple union argument.

Now, we apply the bound on the OGD regret of Proposition~\ref{prop:ogd_full} with $a=m\epsilon_h$, $A=hML^2$, namely that the \textit{good event}
\begin{align*}
	\forall n\leq N,\ \sum_{j=1}^{n-1} \ell_j(z_j) - \ell_j(\bar{z}_n) \leq \frac{C'd h\sigma^2}{\epsilon_h} \log\left(\frac{2dN}{\delta}\right)\log(n)\,,
\end{align*}
holds with probability at least $1-\delta$, for some constant $C'>0$ (in which we hide the dependency on $h, M, L, \alpha$ and $S$ to avoid further cluttering). We assume to be on this event in the rest of the proof, which we combine to the previous events with a union argument, leading to a probability ot at least $1 - (N+1)\delta$.

The crux of the argument is similar to the proof of Lemma~2 in \citep{ding2021efficient} and exploits the strong convexity of the losses $\ell_j$ to relate the online regret to a control on the distance $\lVert \tilde{z}_n - \bar{z}_n\rVert$. By Jensen's inequality, we have
\[\sum_{j=1}^{n} \ell_j(\tilde{z}_n) - \ell_j(\bar{z}_n) \leq \frac{Cd h\sigma^2}{\epsilon_h} \log\left(\frac{2dN}{\delta}\right)\log(n)\,.\]

Strong convexity also implies the following inequality:
\begin{align*}
\ell_j(\tilde{z}_n) - \ell_j(\bar{z}_n) \geq \langle \nabla \ell_j(\bar{z}_n), \tilde{z}_n - \bar{z}_n\rangle + \frac{m\epsilon_h}{2} \lVert \tilde{z}_n - \bar{z}_n\rVert^2_2\,.
\end{align*}
Summing over $j=1, \dots, n$ and exploiting the fact that the sum of gradients vanishes at $\bar{z}_n$, we obtain after some simple algebra:
\[\lVert \tilde{z}_n - \bar{z}_n\rVert^2_2 \leq \frac{2Cd h\sigma^2}{m\epsilon_h^2 n} \log\left(\frac{2dN}{\delta}\right)\log(n)\,.\]

\paragraph{Regret Analysis of LinUCB-OGD} Mirroring the regret proof of LinUCB, we see that we need 
\[\forall t\leq T,\ \Delta(X_t, \bar{\theta}^{\text{OGD}}_n) \leq \gamma_t(X_t)\]
to hold with high probability, for a certain exploration sequence $(\gamma_t)_{t\in\N}$. This amount to controlling the following norm:
\begin{align*}
\lVert \theta^* - \bar{\theta}^{\text{OGD}}_n\rVert_{\bar{H}^{\alpha}_t(\theta^*, \bar{\theta}^{\text{OGD}}_n)} &\leq \lVert \theta^* - \bar{\theta}_t\rVert_{\bar{H}^{\alpha}_t(\bar{\theta}_t, \bar{\theta}^{\text{OGD}}_n)} + \lVert \bar{\theta}^{\text{OGD}}_n - \bar{\theta}_t\rVert_{\bar{H}^{\alpha}_t(\bar{\theta}_t, \bar{\theta}^{\text{OGD}}_n)}\\
&\leq \lVert \theta^* - \bar{\theta}_t\rVert_{\bar{H}^{\alpha}_t(\theta^*, \bar{\theta}^{\text{OGD}}_n)} + \sqrt{M}\lVert \bar{\theta}^{\text{OGD}}_n - \bar{\theta}_t\rVert_{V^{\frac{\alpha}{m}}_t}\\
&\leq \lVert \theta^* - \bar{\theta}_t\rVert_{\bar{H}^{\alpha}_t(\theta^*, \bar{\theta}^{\text{OGD}}_n)} + \sqrt{M\left(L^2 t+\frac{\alpha}{m}\right)}\lVert \bar{\theta}^{\text{OGD}}_n - \bar{\theta}_t\rVert_2\,.
\end{align*}

The first term can be controlled by transportation of local metrics in the same way as in the proof of Theorem~\ref{thm:regret_ucb}, i.e.,
\begin{align*}
\lVert \theta^* - \bar{\theta}_t\rVert_{\bar{H}^{\alpha}_t(\theta^*, \bar{\theta}^{\text{OGD}}_n)} &\leq \sqrt{\kappa}\left(\lVert F^{\alpha}_t(\theta^*) - F^{\alpha}_t(\wh{\theta}_t)\rVert_{H^{\beta}_t(\theta^*)^{-1}} + \lVert  F^{\alpha}_t(\bar{\theta}_t) - F^{\alpha}_t(\wh{\theta})\rVert_{H^{\beta}_t(\bar{\theta}_t)^{-1}}\right)\,,
\end{align*}
and thus this term adds the same contribution to the design of the exploration bonus sequence and to the regret bound. For the second term, we apply the previous bound on $\lVert \tilde{z}_{n} - \bar{z}_{n}\rVert_2 = \lVert \bar{\theta}^{\text{OGD}}_n - \bar{\theta}_{t}\rVert_2$. Combining these two inequalities results in the following control:
\begin{align*}
\lVert \theta^* - \bar{\theta}^{\text{OGD}}_n\rVert_{\bar{H}^{\alpha}_t(\theta^*, \bar{\theta}^{\text{OGD}}_n)} &\leq \underbrace{c^\delta_t}_{\makecell[l]{\text{same as in}\\ \text{the proof of}\\ \text{Theorem~\ref{thm:regret_ucb}}}} +\quad \underbrace{\sqrt{\left(L^2 + \frac{\alpha}{mMt}\right) \left( \frac{2\kappa Cd h^2\sigma^2}{\epsilon_h^2} \log\left(\frac{2dT}{h\delta}\right)\log\left(\frac{t}{h}\right)\right)}}_{\eqqcolon\  c^{\text{OGD}, \delta}_{t, T}\ =\ \cO\left(\sigma L\sqrt{\kappa d}\log T\right)}\,.
\end{align*}

The rest of the proof is now identical to that of Theorem~\ref{thm:regret_ucb_bis_full}, i.e., we use the exploration bonus sequence
\[\gamma^{\text{OGD}}_{t, T}\colon x\in\cX_t \mapsto (c^{\delta}_t + c^{\text{OGD}, \delta}_{t, T}) \lVert x \rVert_{H^{\kappa \alpha}_t(\bar{\theta}^{\text{OGD}}_{\lfloor\!\frac{t\!-\!1}{h} \!\rfloor})^{-1}}\,,\]
and control $\sum\limits_{t=1}^T \lVert X_t \rVert_{H^{\kappa \alpha}_t(\bar{\theta}^{\text{OGD}}_{\lfloor\!\frac{t\!-\!1}{h} \!\rfloor})^{-1}}$ using an elliptical potential lemma (since we operate under Assumption~\ref{ass:6} for the strong convexity of the episodic losses $\ell_j$, we use the strong regret guarantee provided by Lemma~\ref{lem:sto_ell_pot}; note that the high probability event on which this lemma applies is already included in the above events, for a total probability of at least $1-(N+1)\delta$).

$\hfill\blacksquare$
}

\begin{remark}[The importance of strong convexity]
	The key argument behind the proof of Theorem~\ref{thm:regret_ucb-ogd} is that the aggregated loss over an episode $\ell_n\colon \theta\in\Theta \mapsto \sum\limits_{k=1}^{h} \cL(Y_{(n-1)h+k}, \langle \theta, X_{(n-1)h+k}\rangle) \!+\! \frac{\alpha}{2N} \lVert \theta \rVert^2_2$ is $m\epsilon_h$ strongly convex. With simple convexity only, the online regret guarantee of the OGD approximation scales like $\cO(\sqrt{T})$ instead of logarithmically. This would only guarantee $\lVert \bar{\theta}^{\text{OGD}}_n - \bar{\theta}_t\rVert_2 = \cO(\sqrt{t})$, resulting in linear $\cO(T)$ bandit regret after multiplying this term with the contribution of the elliptic potential lemma. Moreover, although $\ell_n$ is always trivially at least $\frac{\alpha}{N}$-strongly convex, it is necessary to ensure non-vanishing strong convexity when $T\rightarrow +\infty$ (we recall that $N=\lceil\frac{T-1}{h}\rceil$). Indeed, substituting $\epsilon_h$ with $\frac{\alpha}{N}$ in the regret bound above gives $\cR_T \leq \cO(\epsilon_h^{-1})=\cO(T)$.
\end{remark}

\paragraph{Scaling of episode length $h$}
As shown in the proof, non-vanishing strong convexity of $\ell_n$ can be deduced from a fixed lower bound on the smallest eigenvalue of the Hessian of $\ell_n$. By Lemma~\ref{lem:smallest_eig}, this holds with high probability provided the episode length $h$ is high enough, which translates to
$h=\lceil \frac{2\epsilon_h}{\rho_{\cX}L^2} + \frac{8}{\rho_{\cX}^2}\log\frac{2}{\delta}\rceil$. Using the typical bound $\rho_{\cX}=\cO(d^{-1})$, we see that $h$ scales like $\cO(d^2)$ in the action dimension. By comparison, the only similar OGD scheme for generalized linear bandit scales like $\cO(d^3)$ \citep[Lemma~2 and Remark~2]{ding2021efficient}, thus suffering to a greater extent from the curse of dimensionality. Note the practical tradeoff on $h$ faced by the agent running Algorithm~\ref{algo::UCB-OGD_convex_risk}: the higher $h$, the more likely it is that the OGD estimator $\bar{\theta}^{\text{OGD}}$ well approximates the true empirical risk minimizer $\bar{\theta}$ (because of stronger convexity of the episodic losses); however, it also means longer episodes and thus less frequent updates of $\bar{\theta}^{\text{OGD}}$, i.e., less learning.

Note that the value of $h$ is derived from a concentration bound that is \textit{uniform} in $h$ (Lemma~\ref{lem:smallest_eig}). However, since $h$ is kept constant throughout the run of the algorithm, a similar, non-uniform result would actually be sufficient (we chose to use Lemma~\ref{lem:smallest_eig} mainly for the sake of convenience, since we already assumed to be on the corresponding good event in order to mirror the regret analysis of Theorem~\ref{thm:regret_ucb_bis_full}). It is actually possible to tighten the lower bound on $h$ using a finer, non-uniform concentration result, which we state below.

\begin{restatable}[Tighter bound on the episode length $h$]{lem}{lem_smallest_eig_tighter}\label{lem:smallest_eig_tighter}
	Under Assumptions~\ref{ass:5} and \ref{ass:6}, let $\epsilon_h>0$, $\delta\in(0, 1)$ and define 
	\[h = \left\lceil\frac{1}{4\rho_{\cX}^2}\left(\sqrt{2(1+\gamma_\delta)\log\left(\frac{2}{\delta}\sqrt{1+\frac{1}{\gamma_\delta}}\right)} + \sqrt{\sqrt{2(1+\gamma_\delta)\log\left(\frac{2}{\delta}\sqrt{1+\frac{1}{\gamma_\delta}}\right)} + \frac{\rho_{\cX}\epsilon_h}{L^2}}\right)^2\right\rceil \,,\]
	where $\gamma_\delta=\frac{-1}{1+W_{-1}(-\frac{\delta^2}{4e})}$ and $W_{-1}$ is the first lower branch of the Lambert W function, i.e., the smallest real solution for $z=[-\frac{1}{e}, 0)$ of the equation $W_{-1}(z)e^{W_{-1}(z)} = z$.
	It holds that
	\[\bP\left(\lambda_{\min}\left(V^0_{h+1}\right) \leq  \epsilon_h\right) \leq \delta\,.\]
\end{restatable}

\proof{
The idea is similar to that of the proof of Lemma~\ref{lem:smallest_eig}, i.e., relate the deviation of the smallest eigenvalue to that of a matrix martingale. The difference lies in the choice of the concentration bound for this martingale.	Fix $\eta>0$, \citet[Corollary S1(a)]{howard2021time} with a normal mixture bound shows that a martingale $\left(M_t\right)_{t\in\bN}$ taking values in a Hilbert space $(\cH, \langle \cdot, \cdot\rangle)$ with uniformly bounded increments $\lVert M_{t+1}-M_{t}\rVert \leq c$ for some $c>0$ and all $t\in\bN$ satisfies
	 
\[\bP\left(\exists t\in\bN,\ \lVert M_t \rVert \geq c \sqrt{2(t+\eta)\log\left(\frac{2}{\delta}\sqrt{1+\frac{t}{\eta}}\right)} \right) \leq \delta\,.\]

Now, we fix $h\in\bN$. Although this bound is uniform in $t\in\bN$, we can make use of the free parameter $\eta$ to optimize it for $t=h$. This procedure is standard (see e.g.,  \citet[Proposition~3]{howard2021time}) and yields $\eta = \gamma_\delta h$ with $\gamma_\delta=\frac{-1}{1+W_{-1}(-\frac{\delta^2}{4e})}$. In particular, we have
\[\bP\left(\lVert M_h \rVert \geq c \sqrt{2h(1+\gamma_\delta) \log\left(\frac{2}{\delta}\sqrt{1+\frac{1}{\gamma_{\delta}}}\right)} \right) \leq \delta\,.\]

Following the steps of Lemma~\ref{lem:smallest_eig}, we have that
\[\bP\left(\lambda_{\min}\left(V^0_{h+1}\right) \leq \epsilon_h\right) \leq \bP\left(\lVert M_h \rVert \geq \rho_{\cX}L^2 h - \epsilon_h \right) \leq \delta\,,\]
with $M_h = V^0_{h+1} - \bE\left[V^0_{h+1}\right]$, which is a martingale with increments bounded by $c=2L^2$. Equating both bounds on $\lVert M_h\rVert$ yields
\[\rho_{\cX}L^2 h - 2L^2\sqrt{2h(1+\gamma_\delta) \log\left(\frac{2}{\delta}\sqrt{1+\frac{1}{\gamma_{\delta}}}\right)} - \epsilon_h = 0\,. \]
This expression is a quadratic equation in $H=\sqrt{h}$. Let $a=\rho_{\cX}L^2$ and $b=L^2\sqrt{2h(1+\gamma_\delta) \log\left(\frac{2}{\delta}\sqrt{1+\frac{1}{\gamma_{\delta}}}\right)}$. Notice that both $a$ and $b$ are positive and that the discriminant of $aH^2 + 2bH - \epsilon_h=0$ is $4b^2+4a\epsilon_h$, which is also positive. The only positive solution is thus given by $\sqrt{h}=H=\frac{b+2\sqrt{b^2+a\epsilon_h}}{2a}$, which concludes the proof.

$\hfill\blacksquare$	
}

The value of $h$ recommended by Lemma~\ref{lem:smallest_eig_tighter} scales with $\rho_{\cX}^{-2}$ at the first order, and thus $h=\cO(d^2)$, just like in Theorem~\ref{thm:regret_ucb_bis_full}. However, it is typically smaller, and thus more practical (allows for more frequent OGD updates with the same theoretical guarantees). We report in Figure~\ref{fig:h_ogd} the numerical values for both expressions of $h$ for typical choices of the parameters $L$, $\epsilon_h$ and $\delta$ as a function the dimension $d$. In practice, even smaller values of $h$ may ensure enough convexity of the episodic losses to observe sublinear regret, which we empirically witnessed in the experiments (see Appendix~\ref{app:xp}). For the practitioner, $h$ may be viewed as a hyperparameter to be tuned manually, potentially on an instance-dependent basis, with Theorem~\ref{thm:regret_ucb_bis_full} and Lemma~\ref{lem:smallest_eig_tighter} giving only worst-case guarantees.

\begin{figure}
	\centering
	\includegraphics[height=0.3\textheight]{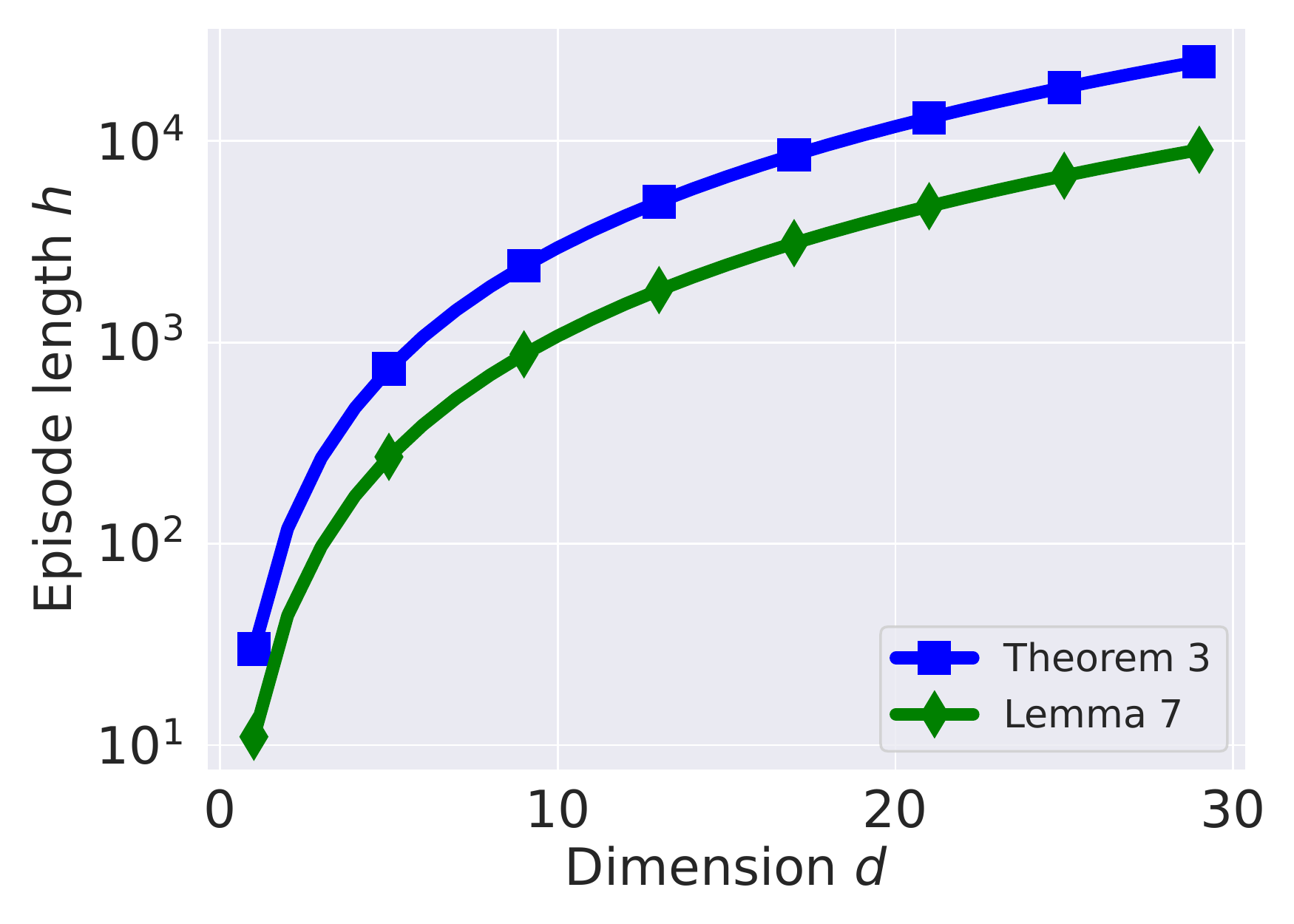}
	\caption{Recommended episode length $h=\lceil \frac{2\epsilon_h}{\rho_{\cX}L^2} + \frac{8}{\rho_{\cX}^2}\log\frac{2}{\delta}\rceil$ for LinUCB-OGD-CR (Algorithm~\ref{algo::UCB-OGD_convex_risk}) according to Theorem~\ref{thm:regret_ucb-ogd} and Lemma~\ref{lem:smallest_eig_tighter} as a function of $d=\rho_{\cX}^{-1}$, for typical values $L=1$,  $\epsilon_h=0.1$ and $\delta=5\%$.}
	\label{fig:h_ogd}
\end{figure}
\section{EXPERIMENTS}\label{app:xp}

We report three simple experiments, two in the expectile setting and one in the entropic risk setting.

\paragraph{Computing Expectiles} We detail two cases of distributions for which expectiles are known. For $p\in(0, 1)$, we denote by $e_p(\nu)$ the $p$-expectile of distribution $\nu$.
\begin{itemize}
	\item If $\nu=\cN\left(0, 1\right)$, then, letting $\phi$ and $\Phi$ be the pdf and cdf of $\nu$ respectively, we obtain after simple calculus and the identity $\phi'(y) = -y\phi(y)$ the following fixed point equation:
	\[e_p(\nu) = \frac{2p\phi(e_p(\nu)) - 1}{(1-2p)\Phi(e_p(\nu)) + p}\,,\]
	from which one can estimate the value of $e_p(\nu)$ using a fast iterative scheme. The general Gaussian case $\nu=\cN\left(\mu, \sigma^2\right)$ is then easily deduced from the relation $e_p(\nu) = \mu + \sigma e_p(\cN\left(0, 1\right))$. Expectile calculations for a few other classical distributions are covered in \citet{philipps2022interpreting}.
	\item If $\nu$ is the so-called expectile based distribution \citep{torossian2020bayesian, arbel2021multivariate} with asymmetric density (with respect to the Lebesgue measure) given by
	\[f_{\mu, \sigma, p}(y) = \frac{\sqrt{2p(1-p)}}{\sigma\sqrt{\pi}(\sqrt{p} + \sqrt{1-p})} \exp\left(-\frac{\lvert p - \ind_{y<\mu}\rvert (y-\mu)^2}{2\sigma^2}\right)\,,\]
	then $e_p(\nu) = \mu$. In other words, these distributions offer a family parametrized directly by their expectile, generalizing the family of Gaussian distributions parametrized by their mean (for a given variance).
\end{itemize}

We recall that the $p$-expectile can be elicited by the convex potential $\psi(z) = \lvert p-\ind_{z<0}\rvert z^2$. The second derivative of this potential is given by $\psi''(z) = 2(1-p)\ind_{z<0} + 2p\ind_{z>0}$, which is bounded between $2p$ and $2(1-p)$. In particular, Assumption~\ref{ass:loss_curvature} holds with conditioning $\kappa=\frac{M}{m}=\frac{1-p}{p}$ if $p\leq\frac{1}{2}$ and $\kappa=\frac{M}{m}=\frac{p}{1-p}$ otherwise. Note that the two classes of distributions considered above are Gaussian or log-concave, which fits the scope of the supermartingale control of Lemma~\ref{lem:supermart}. 

\paragraph{Computing Entropic Risk} For a distribution $\nu$, the entropic risk at level $\gamma>0$ takes the form $\rho_{\gamma}(\nu) = \frac{1}{\gamma}\!\log \bE_{Y\sim\nu}\!\left[e^{\gamma Y}\right]$ and corresponds to the loss $\cL\colon (y, \xi)\mapsto \xi + \frac{1}{\gamma}(e^{\gamma(y-\xi)}-1)$. Derivatives of this loss satisfy the following identities, where $\partial$ represents the differentiation operator with respect to the second coordinate $\xi$:
\begin{align*}
\partial \cL(y, \xi) &= 1 - e^{\gamma(y-\xi)},\\
\partial^2 \cL(y, \xi) &= \gamma e^{\gamma(y-\xi)},
\end{align*}
and is thus in particular strictly convex.

For a Bernoulli-like distribution $\nu = p\delta_a + (1-p)\delta_b$, with $p\in(0, 1)$, $a, b\in \R$, the entropic risk takes the simple form $\rho_{\gamma}(\nu) = \frac{1}{\gamma}\log\left(p e^{\gamma a} + (1 - p)e^{\gamma b}\right)$. If $\nu$ has a bounded support with diameter $\cD$, then it is clear that the Hessian of the loss is controlled by $m=\gamma e^{-\gamma \cD}\leq \partial^2 \cL \leq \gamma e^{\gamma \cD}=M$, and therefore the conditioning number of the loss $\kappa$ can be bounded by $e^{2\gamma \cD}$. Finally, $\nu$ being bounded also fits the scope of the supermartingale control of Lemma~\ref{lem:supermart}.

\paragraph{General Case} If a density $p$ and a loss function $\cL$ are known, one may resort to numerical integration to approximate the following quantity up to arbitrary precision:
\[\bE_{Y\sim\nu}\left[ \cL(Y, \xi) \right] = \int \cL(y, \xi)p(y)dy \approx \sum_{i} w_i \cL(y_i, \xi)p(y_i),\]
where the weights $(w_i)$ and knots $(y_i)$ depend on the approximation routine. Then, one may simply run a minimization algorithm on the function $\xi\mapsto \sum_i w_i \cL(y_i, \xi)p(y_i)$ to estimate $\rho_{\cL}(\nu)$.

\paragraph{Experiment 1: Multi-armed Gaussian Bandit with Expectile Noise}
We considered $K=2$ Gaussian arms with expectiles at level $p=10\%$ equal to $1$ and $0$ respectively. This bandit can be represented by constant orthonormal actions $\cX_t = \lbrace [1\ 0]^\top,\ [0\ 1]^\top\rbrace$, parameter $\theta^* = [1\ 0]^\top$ and noise distributions $\cN\left(\mu_k, \sigma_k^2\right)$, with $\mu_k$ and $\sigma_k$ chosen such that the expectile of the corresponding noise is zero for $k\in\left\{1, 2\right\}$. This can be achieved with e.g., $\mu_1\approx 0.44$, $\sigma_1=0.5$ and $\mu_2\approx 2.62$, $\sigma_2=3$, which was the setup for this experiment. Note that for a given expectile level $p\in(0, 1)$ and standard deviation $\sigma$, finding the unique mean $\mu$ such that $\cN\left(\mu, \sigma^2\right)$ has zero $p$-expectile can be easily done via a numerical root search, using the formula for Gaussian expectiles described above.

The optimal arm with respect to the expectile criterion is the first one by definition. However, the expectations of these arms are in reversed order, making the second one optimal with respect to the mean criterion. 

\paragraph{Experiment 2: Linear Bandit with Expectile Asymmetric Noise}

We considered a second example with non-Gaussian noise and non-orthogonal features. We defined the action set at time $t$ by $\cX_t = \lbrace X^1_t, X^2_t\rbrace\subset \R^3$ where:
\begin{itemize}
	\item $X^1_t=\frac{Z^1_t}{\lVert Z^1_t\rVert}$ with $Z^1_t\sim\cN([1\ 0\ 0]^\top, \sigma_x I_3)$,
	\item $X^2_t=\frac{Z^2_t}{\lVert Z^2_t\rVert}$ with $Z^1_t\sim\cN([0\ 1\ 0]^\top, \sigma_x I_3)$,
	\item We set the action noise to an arbitrary value $\sigma_x=0.1$.
	\item $(Z^1_t, Z^2_t)_{t\in\N}$ are all independent random variables.
\end{itemize} 

This construction results in bounded, anisotropic actions. We chose $\theta^* = [0.9\ 0\ 1]^\top$, so that $\langle \theta^*, X^1_t\rangle$ is likely higher than $\langle \theta^*, X^2_t\rangle$, thus favoring $X_t = X^1_t$ in the expectile model $Y_t = \langle \theta^*, X_t\rangle + \eta_t$. To model the zero $p$-expectile noise $\eta_t$ with $p=10\%$, we used the expectile based distribution presented above with $\mu_1=\mu_2=0$ and $\sigma_1=0.5$ if action $X^1_t$ is played, and $\sigma_2=1.5$ otherwise, resulting in different mean noise $\bE[\eta_t\lvert \cF_t] \approx 1.8$ and $\bE[\eta_t\lvert \cF_t] \approx 3.3$ respectively. As in the previous example, this setting was designed to deceive the mean criterion by inverting the order of optimal actions. 

\paragraph{Experiment 3: Multi-armed Bernoulli Bandit with Entropic Risk Noise}

The last experiment consisted of $K=2$ Bernoulli-like arms $\nu_1=\frac{1}{2}\delta_{1} + \frac{1}{2}\delta_{-1}$ and $\nu_2=\frac{1}{4}\delta_{2} + \frac{3}{4}\delta_{-2}$, which corresponds to means $\mu_1=0$, $\mu_2=-1$ and entropic risk $\rho_\gamma(\nu_1)\approx 0.43$ and $\rho_\gamma(\nu_2)\approx 0.67$ at level $\gamma=1$. Again, this setting was designed so that the best optimal arm is different under the mean and entropic risk criteria.

\paragraph{Results}
On each of the three settings, we ran an instance of Algorithm~\ref{algo::UCB_convex_risk}, i.e., LinUCB (convex risk), and Algorithm~\ref{algo::UCB-OGD_convex_risk}, i.e., LincUCB-OGD (convex risk). We also ran a standard LinUCB algorithm for the mean criterion \citep{abbasi_yadkori2011}. Hyperparameters $m, M$ and $\kappa$ were tuned according to the analysis above. Regularization was fixed at $\lambda=0.1$. As is customary in bandit experiments, the parameter $\sigma$, which in the formal analysis is derived from the supermartingale control of the noise, was considered a degree of freedom to control the amount of exploration; we arbitrarily fixed it at $\sigma=0.1$ in experiments 1 and 2 and at $\sigma=1$ in experiment 3. For the LinUCB-OGD variant, the step size for the OGD scheme was set to $\epsilon_n = 0.1/n$, following the linear decay suggested by Proposition~\ref{prop:ogd}, and the frequency of OGD update to $h=5$. In addition, all algorithms went through an initial warmup phase where each arm was played $5$ times, in order to ensure better stability of the initial estimations of $\theta$.

In all three examples, the mean criterion algorithm was deceived and accumulated linear expectile and entropic risk regret, while both risk-aware algorithms exhibited sublinear trends. Interestingly, the LinUCB-OGD variant showed higher regrets due to the approximate minimization of the loss criterion by OGD, but remained below the mean criterion LinUCB benchmark. Figure~\ref{fig:xp_expectile2} reproduces the results of each experiments across 500 independent replications. Finally, average runtimes for each algorithm are reported in Table~\ref{table:runtime1}. Calculations were performed on a distributed infrastructure comprised of 80 CPUs. While the values themselves are not indicative, as they would vary on a different system, their relative magnitudes illustrate the computational gain of the OGD scheme over solving the empirical risk minimization problem at each step as required in LinUCB (convex risk). Note also that the standard LinUCB with mean criterion is faster due to the sequential nature of the ridge regression estimator. Indeed, this procedure involves inverting at each step a $d\times d$ matrix subject to rank one updates, which can be calculated efficiently via the Sherman-Morrison formula. By contrast, other convex losses than the one derived from the quadratic potential loose this sequential form and require solving the corresponding regression problem from scratch at each time step.

\begin{figure}
	\centering
	\begin{subfigure}[b]{0.49\textwidth}
		\centering
		\includegraphics[height=0.37\textheight]{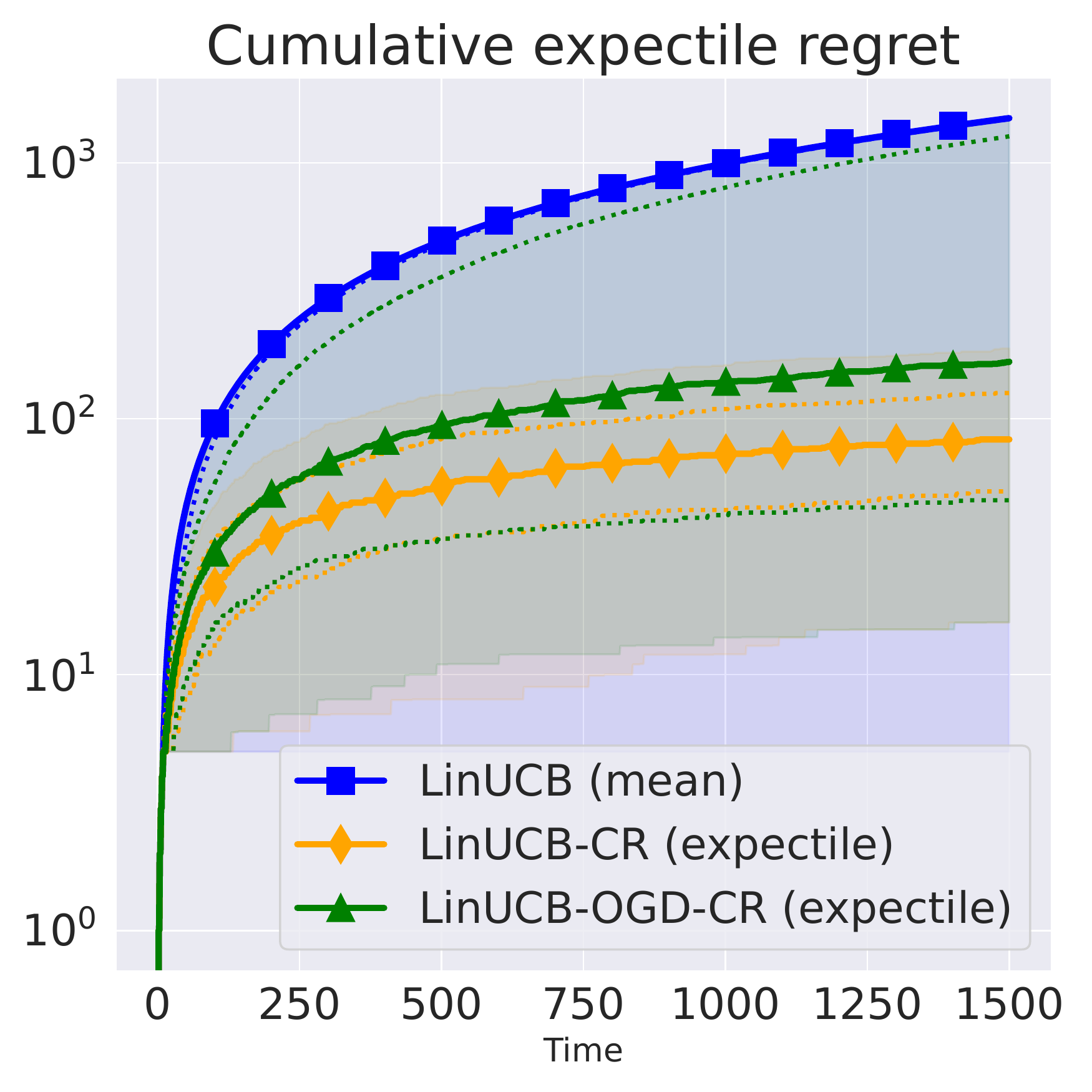}
	\end{subfigure}
	\begin{subfigure}[b]{0.49\textwidth}
		\centering
		\includegraphics[height=0.37\textheight]{figures/expectile_linucb.pdf}
	\end{subfigure}\\
	\begin{subfigure}[b]{0.49\textwidth}
		\centering
		\includegraphics[height=0.37\textheight]{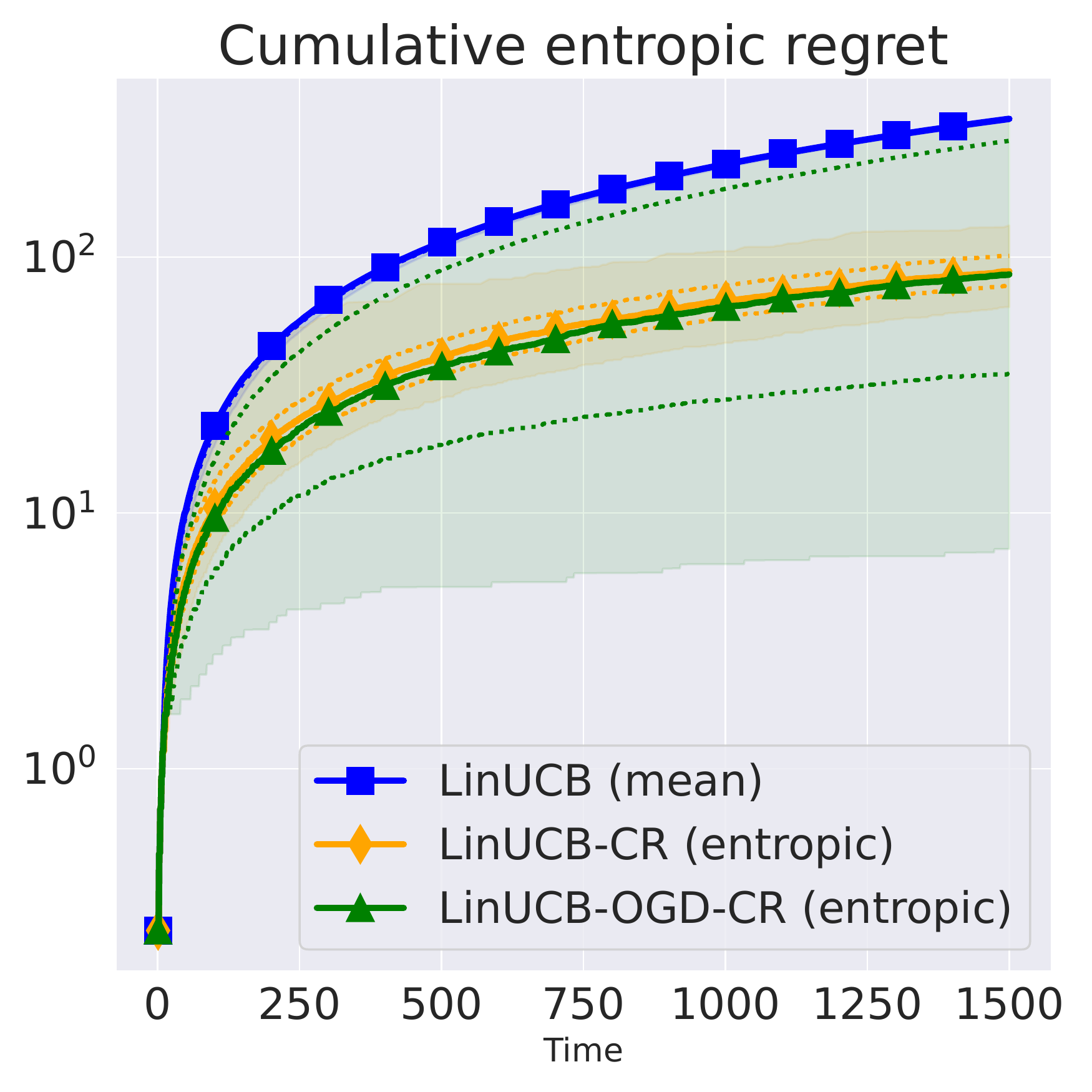}
	\end{subfigure}
	\caption{Left: two-armed Gaussian expectile bandit. Center: two-armed linear expectile bandit with $\bR^3$ contexts and expectile-based asymmetric noises. Right: two-armed Bernoulli entropic risk bandit. Thick lines denote median cumulative regret over 500 independent replications. Dotted lines denote the 25 and 75 regret percentiles. Shaded areas denote the 5 and 95 percentiles.}
	\label{fig:xp_expectile2}
\end{figure}

\begin{small}
	\begin{table}[!ht]
		\caption{Runtimes for the Classical LinUCB and Algorithms~\ref{algo::UCB_convex_risk} (LinUCB for Convex Risk) and \ref{algo::UCB-OGD_convex_risk} (LinUCB-OGD for Convex Risk) in Each Experiments. Runtimes are Reported in Seconds as Mean $\pm$ Standard Deviation, Estimated Across 500 Independent Replications with Time Horizon $T=1500$.}
		\label{table:runtime1}
		\centering
		\begin{tabular}{llll}
			\toprule
			Algorithm     & Experiment 1 & Experiment 2 & Experiment 3 \\
			\midrule
			LinUCB (mean) & $0.4 \pm 0.0$ & $37.2 \pm 4.9$ & $0.6 \pm 0.0$  \\[1.0ex]
			LinUCB-CR (convex risk) & $231.0 \pm 21.7$ & $814.8 \pm 88.3$ & $519.1 \pm 33.3$ \\[1.0ex]
			LinUCB-OGD-CR (convex risk) & $20.4 \pm 3.9$& $60.2 \pm 12.0$ & $25.7 \pm 4.9$ \\[1.0ex]
			\bottomrule
		\end{tabular}
		\vskip -5mm
	\end{table}
\end{small}

\end{document}